\documentclass[10pt,twocolumn,letterpaper]{article}

\usepackage[pagenumbers]{utils/cvpr} 

\usepackage[dvipsnames]{xcolor}

\usepackage{times}
\usepackage{epsfig}
\usepackage{graphicx}
\usepackage{amsmath}
\usepackage{amssymb}
\usepackage{multirow}
\usepackage{booktabs}
\usepackage{rotating}
\usepackage{overpic}
\usepackage{color}
\usepackage{colortbl}

\usepackage{algorithmic}
\usepackage[linesnumbered,boxed]{algorithm2e}

\def\ie{\emph{i.e.}}
\def\eg{\emph{e.g.}}
\def\etc{\emph{etc}}
\def\etal{{\em et al.~}}

\newcommand{\secref}[1]{Sec.~\ref{#1}}

\makeatletter
\newcommand{\thickhline}{%
    \noalign {\ifnum 0=`}\fi \hrule height 1pt
    \futurelet \reserved@a \@xhline
}

\def\metricsDIS{
    &$F_{\beta}^{x}\uparrow$
    &$F_{\beta}^{\omega}\uparrow$
    &$\mathcal{M}\downarrow$
    &$S_m\uparrow$
    &$E_{\phi}^{m}\uparrow$
    &$HCE_{\gamma}\downarrow$
}
\def\metricsCOD{
    &$S_m\uparrow$
    &$F_{\beta}^{\omega}\uparrow$
    &$F_{\beta}^{m}\uparrow$
    &$E_{\phi}^{m}\uparrow$
    &$E_{\phi}^{x}\uparrow$
    &$\mathcal{M}\downarrow$
}
\def\metricsSODHR{
    &$S_m\uparrow$
    &$F_{\beta}^{x}\uparrow$
    &$E_{\phi}^{m}\uparrow$
    &$\mathcal{M}\downarrow$
}
\def\metricsSODLR{
    &$S_m\uparrow$
    &$F_{\beta}^{x}\uparrow$
    &$E_{\phi}^{m}\uparrow$
    &$\mathcal{M}\downarrow$
}

\def\ourmodel{\textit{BiRefNet}}

\definecolor{myGray}{gray}{.92}
\definecolor{myRed}{RGB}{219, 68, 55}
\definecolor{myGreen}{RGB}{15, 157, 88}
\definecolor{myBlue}{RGB}{66, 133, 244}

\newcommand\blfootnote[1]{%
\begingroup
\renewcommand\thefootnote{}\footnote{#1}%
\addtocounter{footnote}{-1}%
\endgroup
}

\definecolor{cvprblue}{rgb}{0.21,0.49,0.74}
\usepackage[pagebackref,breaklinks,colorlinks,citecolor=cvprblue]{hyperref}

\usepackage[capitalize]{cleveref}
\crefname{section}{Sec.}{Secs.}
\Crefname{section}{Section}{Sections}
\Crefname{table}{Table}{Tables}
\crefname{table}{Tab.}{Tabs.}

\def\paperID{3667} 
\def\confName{CVPR}
\def\confYear{2024}

\begin{document}


\title{Bilateral Reference for High-Resolution Dichotomous Image Segmentation}

\author{
Peng Zheng$^{1,4,5,6\dagger}$ ~~Dehong Gao$^{2}$ ~~Deng-Ping Fan$^{1\ast}$ ~~Li Liu$^{3}$ \\
Jorma Laaksonen$^{4}$ ~~Wanli Ouyang$^{5}$ ~~Nicu Sebe$^{6}$~~\\
$^1$ Nankai University, 
$^2$ Northwest Polytechnical University, \\ 
$^3$ National University of Defense Technology,
$^4$ Aalto University, \\
$^5$ Shanghai AI Laboratory, 
$^6$ University of Trento\\
{\tt\small zhengpeng0108@gmail.com}
}

\twocolumn[{%
  \renewcommand\twocolumn[1][t!]{#1}
  \maketitle
    \begin{center}
        \begin{center}
            \begin{overpic}[width=.98\linewidth]{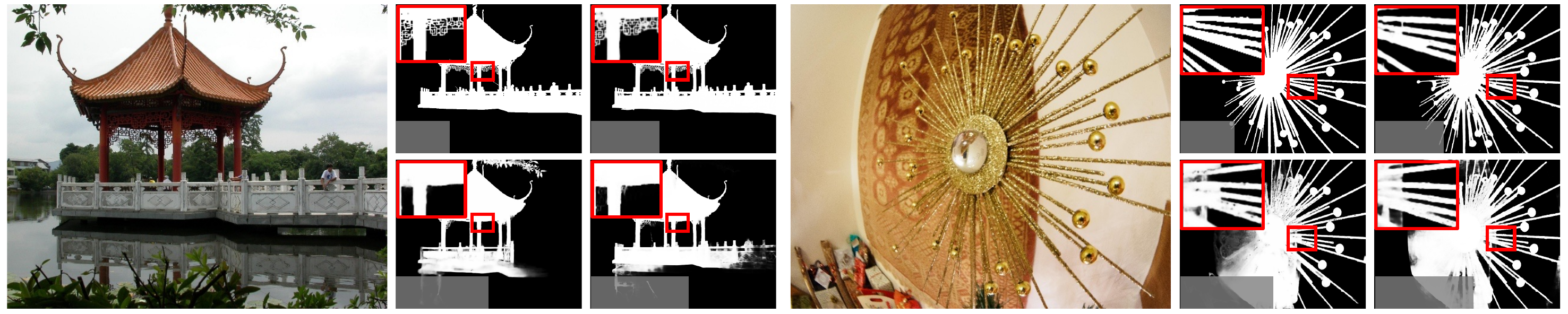}
                \put(25.5,10.5){\textcolor{orange}{GT}}
                \put(37.85,10.5){\textcolor{orange}{Ours}}
                \put(25.5,0.5){\textcolor{orange}{IS-Net}}
                \put(37.85,0.5){\textcolor{orange}{UDUN}}
                
                \put(75.5,10.5){\textcolor{orange}{GT}}
                \put(87.85,10.5){\textcolor{orange}{Ours}}
                \put(75.5,0.5){\textcolor{orange}{IS-Net}}
                \put(87.85,0.5){\textcolor{orange}{UDUN}}
            
            \end{overpic}
        \end{center}
        \captionof{figure}{
        \textbf{Visual comparison between the results of our proposed~\ourmodel{} and the latest state-of-the-art methods} (\eg{}, IS-Net~\cite{DIS5K} and UDUN~\cite{UDUN}) for high-resolution dichotomous image segmentation (DIS). Details of segmentation are zoomed in for better display.
        }
        \label{fig:fig1}
    \end{center}
}]
\blfootnote{$\dagger$ Peng finished the majority of this work when he was a visiting scholar at Nankai University.}
\blfootnote{$\ast$ Corresponding author (dengpfan@gmail.com).}

\maketitle

\vspace{-3mm}
\begin{abstract}
    We introduce a novel bilateral reference framework (\ourmodel{}) for high-resolution dichotomous image segmentation (DIS). 
    It comprises two essential components: the localization module (LM) and the reconstruction module (RM) with our proposed bilateral reference (BiRef). LM aids in object localization using global semantic information. Within the RM, we utilize BiRef for the reconstruction process, where hierarchical patches of images provide the source reference, and gradient maps serve as the target reference. These components collaborate to generate the final predicted maps. We also introduce auxiliary gradient supervision to enhance the focus on regions with finer details. In addition, we outline practical training strategies tailored for DIS to improve map quality and the training process. 
    To validate the general applicability of our approach, we conduct extensive experiments on four tasks to evince that \ourmodel{} exhibits remarkable performance, outperforming task-specific cutting-edge methods across all benchmarks. Our codes are publicly available at \url{https://github.com/ZhengPeng7/BiRefNet}.
\end{abstract}


\section{Introduction}
\label{sec:intro}

\begin{figure*}[t!]
\centering
    \begin{overpic}[width=.98\linewidth]{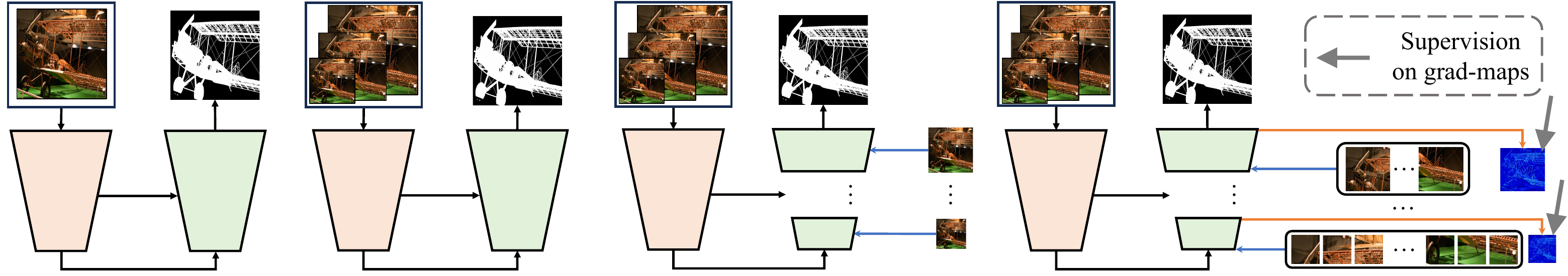}
        \put(7.8,-1.5){(a)}
        \put(27,-1.5){(b)}
        \put(47,-1.5){(c)}
        \put(71,-1.5){(d)}
        
        \put(2.4,4.5){Enc}
        \put(12,4.5){Dec}
        \put(21.5,4.5){Enc}
        \put(31.2,4.5){Dec}
        \put(41.5,4.5){Enc}
        \put(50.5,4.5){Dec}
        \put(66,4.5){Enc}
        \put(75,4.5){Dec}
    \end{overpic}
    \caption{\textbf{Comparison between our proposed~\ourmodel{} and other existing methods for HR segmentation tasks.} (a) Common framework~\cite{UNet}; (b) Image pyramid as input~\cite{image_pyramid_2,zhao2018icnet}; (c) Scaled images as inward reference~\cite{InSPyReNet,PENet}; (d) \ourmodel{}: patches of original images at original scales as inward reference and gradient priors as outward reference. Enc = encoder, Dec = decoder.}
    \label{fig:comp_other_methods}
\end{figure*}
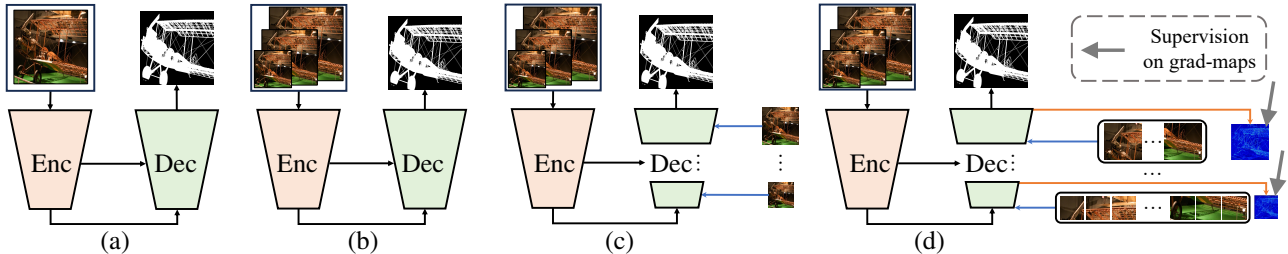

With the advancement in high-resolution image acquisition, image segmentation technology has evolved from traditional coarse localization to achieving high-precision object segmentation. This task, whether it involves salient~\cite{fan2022salient} or concealed object detection~\cite{fan2023advances,SINet_v2}, is referred to as high-resolution dichotomous image segmentation (DIS)~\cite{DIS5K} and has attracted widespread attention and use in the industry, \eg, by Samsung, Adobe, and Disney.

For the new DIS task, recent works have considered strategies such as intermediate supervision~\cite{DIS5K}, frequency prior~\cite{FP-DIS}, and unite-divide-unite~\cite{UDUN}, and have achieved favorable results.
Essentially, they either split the supervision~\cite{UDUN,DIS5K} at the feature-level or introduce an additional prior~\cite{FP-DIS} to enhance feature extraction. These strategies are, however, still insufficient to capture very fine features (see \cref{fig:fig1}). 
Based on our observations, we found that fine and non-salient features in image objects can be well reflected by obtaining gradient features through derivative operations on the original image. 
In addition, when certain positions exhibit high similarity in color and texture to the background, the gradient features are probably too weak. For such cases, we further introduce ground-truth (GT) features for side supervision, allowing the framework to learn the characteristics of these positions.
We name the incorporation of the image reference and the introduction of both the gradient and GT references as \emph{bilateral reference}.

We propose a novel progressive bilateral reference network~\ourmodel{} to handle the high-resolution DIS task with separate localization and reconstruction modules.
For the localization module, we extract hierarchical features from vision transformer backbone, which are combined and squeezed to obtain corase predictions in low resolution in deep layers.
For the reconstruction module, we further design the inward and outward references as bilateral references (BiRef), in which the source image and the gradient map are fed into the decoder at different stages.
Instead of resizing the original images to lower-resolution versions to ensure consistency with decoding features at each stage~\cite{InSPyReNet,PENet}, we keep the original resolution for intact detail features in inward reference and adaptively crop them into patches for compatibility with decoding features.
In addition, we investigate and summarize practical strategies for the training on high-resolution (HR) data, including long training and region-level loss for better segmentation in parts of fine details, and multi-stage supervision to accelerate learning of them.

Our main contributions are summarized as follows:

\begin{enumerate}\setlength\itemsep{0.3em}
    \item{} 
    We present a \textbf{bilateral reference network} (\ourmodel{}), which is a simple yet strong baseline to perform high-quality dichotomous image segmentation.

    \item{} 
    We propose a \textbf{bilateral reference module}, which consists of an inward reference with source image guidance and an outward reference with gradient supervision. It shows great efficacy in the reconstruction of the predicted HR results.

    \item{}
    We explore and summarize various \textbf{practical strategies tailored for DIS} to easily improve performance, prediction quality, and convergence acceleration.

    \item{} 
    The proposed~\ourmodel{} shows its excellent performance and strong generalization capabilities to achieve \textbf{state-of-the-art} performance on not only the \textbf{DIS5K} task but also on \textbf{HRSOD} and \textbf{COD} with \textbf{6.8\%}, \textbf{2.0\%}, and \textbf{5.6\%} average $S_m$~\cite{Smeasure} improvements, respectively.

\end{enumerate}


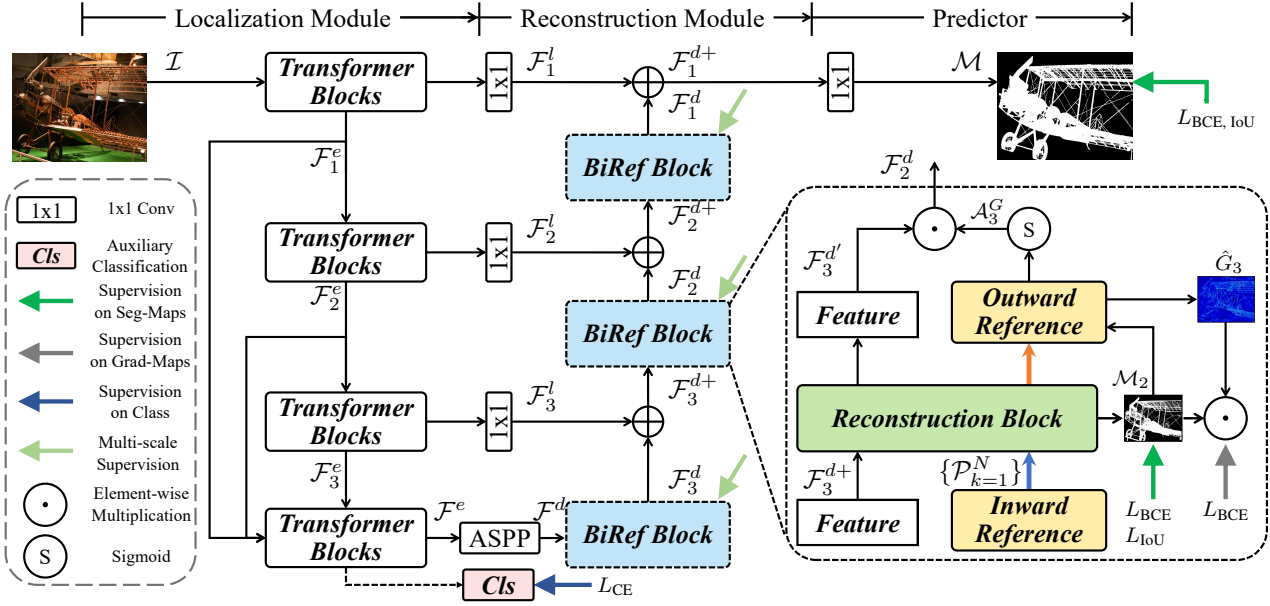
\begin{figure*}[t!]
    \centering
    \begin{overpic}[width=.98\linewidth]{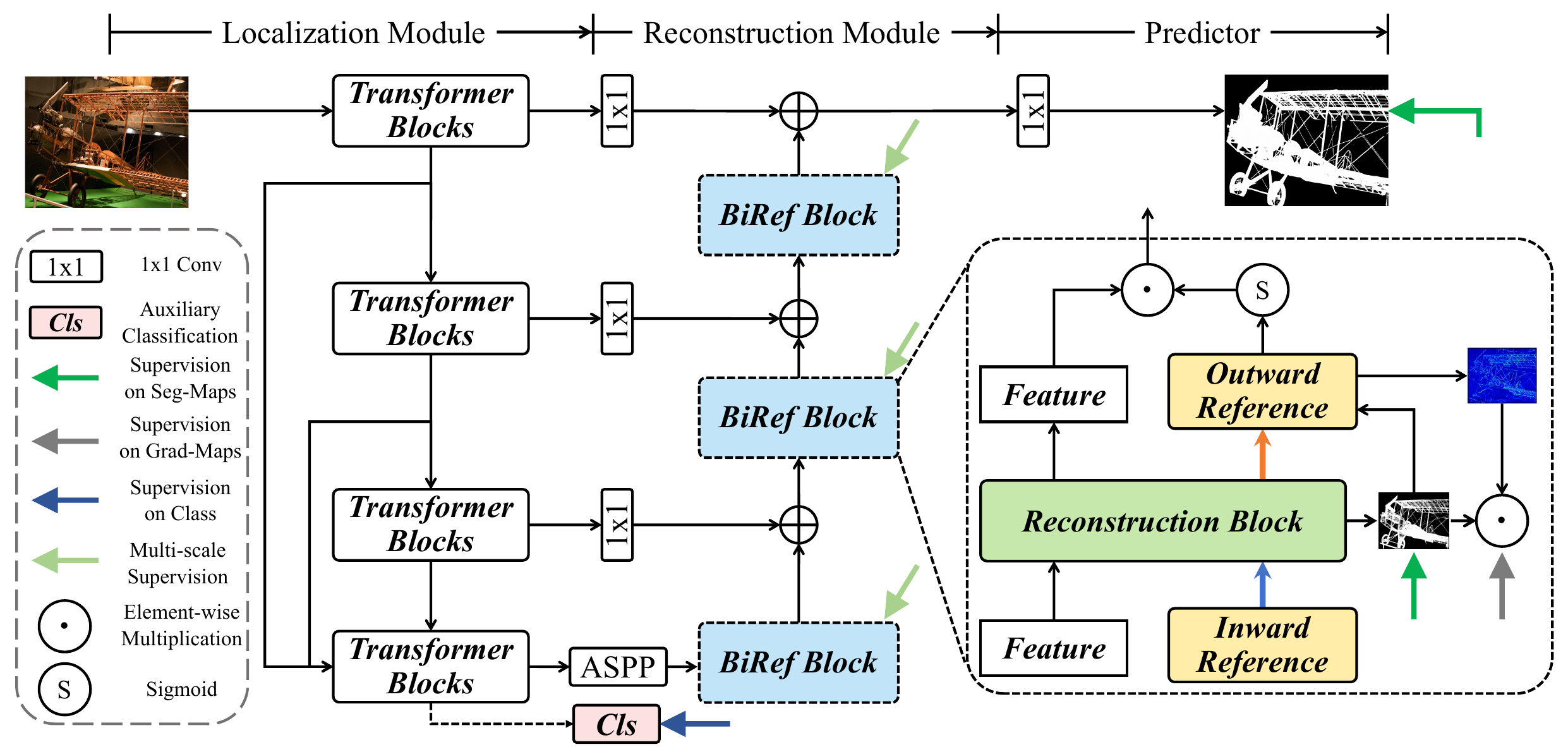}
        \put(13,41.5){\normalsize{} {$\mathcal{I}$}}
        \put(74,41.5){\normalsize{} {$\mathcal{M}$}}

        \put(24.2,34.0){\normalsize{} {$\mathcal{F}^{e}_1$}}
        \put(24.2,23.3){\normalsize{} {$\mathcal{F}^{e}_2$}}
        \put(24.2,9.5){\normalsize{} {$\mathcal{F}^{e}_3$}}
        \put(52,38.2){\normalsize{} {$\mathcal{F}^{d}_1$}}
        \put(52,24.5){\normalsize{} {$\mathcal{F}^{d}_2$}}
        \put(52,9.0){\normalsize{} {$\mathcal{F}^{d}_3$}}
        
        \put(33.8,6.5){\normalsize{} {$\mathcal{F}^e$}}
        \put(41.5,6.5){\normalsize{} {$\mathcal{F}^d$}}

        \put(52,41.5){\normalsize{} {$\mathcal{F}^{d+}_1$}}
        \put(52,29.5){\normalsize{} {$\mathcal{F}^{d+}_2$}}
        \put(52,16){\normalsize{} {$\mathcal{F}^{d+}_3$}}

        \put(41,41.5){\normalsize{} {$\mathcal{F}^{l}_1$}}
        \put(41,28.5){\normalsize{} {$\mathcal{F}^{l}_2$}}
        \put(41,15.5){\normalsize{} {$\mathcal{F}^{l}_3$}}

        \put(68.5,33.5){\normalsize{} {$\mathcal{F}^{d}_2$}}
        \put(62.5,26){\normalsize{} {$\mathcal{F}^{d'}_3$}}
        \put(62.5,9){\normalsize{} {$\mathcal{F}^{d+}_3$}}
        \put(73,9.7){\normalsize{} {$\{\mathcal{P}_{k=1}^N\}$}}

        \put(76,30){\footnotesize {$\mathcal{A}_{3}^{G}$}}
        \put(87,17){\footnotesize {$\mathcal{M}_{2}$}}
        \put(95,25.9){\footnotesize {$\hat{G}_{3}$}}

        \put(47,0.9){\footnotesize {${L}_{\text{CE}}$}}
        \put(92,37.3){\footnotesize {${L}_{\text{BCE, IoU}}$}}
        \put(88,6.8){\footnotesize {${L}_{\text{BCE}}$}}
        \put(88,4.8){\footnotesize {${L}_{\text{IoU}}$}}
        \put(94,6.8){\footnotesize {${L}_{\text{BCE}}$}}
    \end{overpic}
    \caption{
        \textbf{Pipeline of the proposed bilateral reference Network (\ourmodel{}).} 
        \ourmodel{} mainly consists of the localization module (LM) and the reconstruction module (RM) with bilateral reference (BiRef) blocks.
        Please refer to \secref{sec:overview} for details.
    }
    \label{fig:model}
\end{figure*}

\section{Related Works}
\label{sec:related_work}
\subsection{High-Resolution Class-agnostic Segmentation}
\label{sec:HR_seg}

High-resolution class-agnostic segmentation has been a typical computer vision objective for decades, and many related tasks have been proposed and attracted much attention, such as dichotomous image segmentation (DIS)~\cite{DIS5K}, high-resolution salient object detection (HRSOD)~\cite{HRSOD}, and concealed object detection (COD)~\cite{SINet_v2}.
To provide standard HRSOD benchmarks, several typical HRSOD datasets (\eg{}, HRSOD~\cite{HRSOD}, UHRSD~\cite{PGNet}, HRS10K~\cite{RMFormer}) and numerous approaches~\cite{InSPyReNet,DHQSOD,HRSOD,PGNet} have been proposed. Zeng~\etal{}\cite{HRSOD} employed a global-local fusion of the multi-scale input in their network. Xie~\etal{}\cite{PGNet} used cross-model grafting modules to process images at different scales from multiple backbones (lightweight~\cite{resnet} and heavy~\cite{swin_v1}). Pyramid blending was also used in~\cite{InSPyReNet} for a lower computational cost.
Concealed objects are difficult to locate due to similar-looking surrounding distractors~\cite{SINet_v1}. Therefore, image priors, such as frequency~\cite{FDNet}, boundary~\cite{BGNet}, gradient~\cite{DGNet},~\etc{}, are used as auxiliary guidance to train COD models. Furthermore, a higher resolution has been found beneficial for detecting targets~\cite{DGNet,FSPNet,HitNet}. To produce more precise and fine-detail segmentation results, Yin~\etal{}\cite{CamoFormer} employed progressive refinement with masked separable attention. Li~\etal{}\cite{PENet} incorporated the original images at different scales to aid in the refining process.

High-resolution DIS is a newly proposed task that focuses more on the complex slender structure of target objects in high-resolution images, making it even more challenging. Qin~\etal{}\cite{DIS5K} proposed the DIS5K dataset and IS-Net with intermediate supervision to alleviate the loss of fine areas. 
In addition, Zhou~\etal{}\cite{FP-DIS} embedded a frequency prior to their DIS network to capture more details. Pei~\etal{}\cite{UDUN} applied a label decoupling strategy~\cite{LDF} to the DIS task and achieved competitive segmentation performance in the boundary areas of objects. Yu~\etal{}~\cite{dis_mvanet} used patches of HR images to accelerate the training in a more memory-efficient way.
Unlike previous models that used compressed/resized images to enhance HR segmentation, we utilized intact HR images as supplementary information for better predictions in high resolution.

\subsection{Progressive Refinement in Segmentation}
\label{sec:progressive_refinement_in_segmentation}

In the image matting task, trimaps have been used as a pre-positioning technique for more precise segmentation results~\cite{SOD4ImageMatting1,matting_1}. In~\cref{fig:comp_other_methods}, we illustrate different approaches of relevant networks and compare the differences~\cite{UNet,zhao2017pyramid,DIS5K,FP-DIS,UDUN}.
Many approaches have been proposed based on the progressive refinement strategy. Yu~\etal{}\cite{prog_ref_1} used the predicted LR alpha matte as a guide for refined HR maps. In BASNet~\cite{BASNet}, the initial results are revised with an additional refiner network. The CRM~\cite{CRM} continuously aligns the feature map with the refinement target to aggregate detailed features. In ICNet~\cite{zhao2018icnet}, the original images are also downscaled and added to the decoder output at different stages for refinement. In ICEG~\cite{cod_generative_augmentation}, the generator and the detector iteratively evolve through interaction to obtain better COD segmentation results.
In addition to images and GT, auxiliary information is also used in existing methods. For example, Tang~\etal{}\cite{refine_aux_boundary} cropped patches on the boundary to further refine them. In the LapSRN network~\cite{SR_refine_1,SR_refine_2} for image super-resolution, Laplacian pyramids are also generated to help with image reconstruction at higher resolution. 
Although these methods successfully employed refinement to achieve better results, models are not guided to focus on certain areas, which is a problem in DIS. Therefore, we introduce gradient supervision in our outward reference to guide features sensitive to areas with richer fine details.


\section{Methodology}
\label{sec:methodology}

\subsection{Overview}
\label{sec:overview}
As shown in~\cref{fig:comp_other_methods}(d), our proposed~\ourmodel{} is different from the previous DIS methods. On the one hand, our~\ourmodel{} explicitly decomposes the DIS task on HR data into two modules,~\ie{}, a localization module (LM) and a reconstruction module (RM). On the other hand, instead of directly adding the source images~\cite{image_pyramid_2} or the priors~\cite{FP-DIS} to the input, \ourmodel{} employs our proposed bilateral reference in the RM, making full use of the source images at the original scales and the gradient priors. The complete framework of our~\ourmodel{} is illustrated in~\cref{fig:model}.

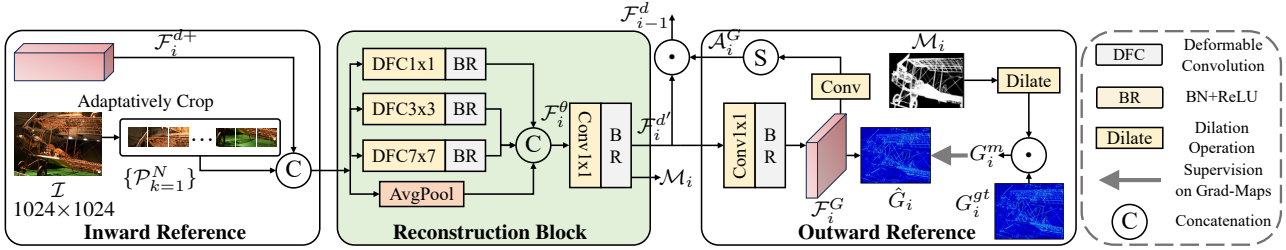
\begin{figure*}[t!]
    \centering
    \begin{overpic}[width=.98\linewidth]{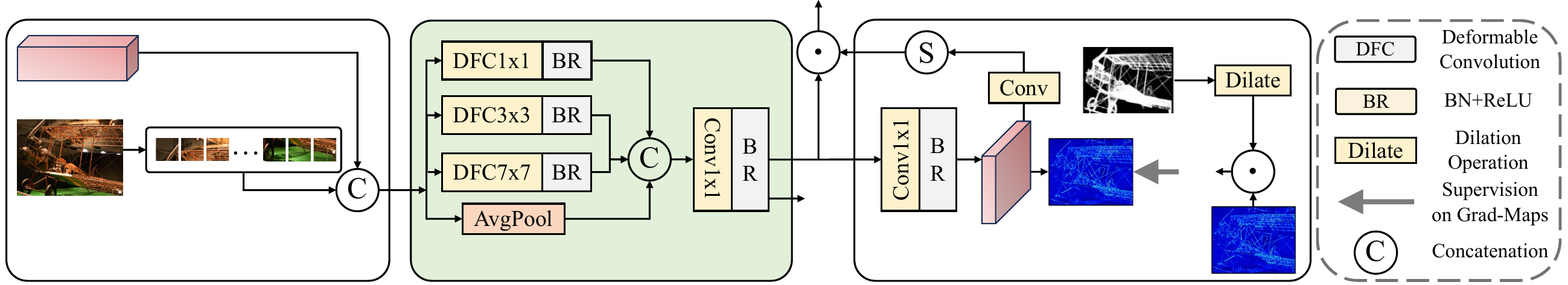}
        \put(4,4){\footnotesize {$\mathcal{I}$}}
        \put(1,2.5){\footnotesize {1024$\times$1024}}
        \put(6,11){\scriptsize {Adaptatively Crop}}
        \put(9.5,5){\footnotesize {$\{\mathcal{P}_{k=1}^N\}$}}
        \put(12,15.5){\footnotesize {$\mathcal{F}_i^{d+}$}}
        \put(42,10){\footnotesize {$\mathcal{F}_{i}^{\theta}$}}
        \put(49.5,9){\footnotesize {$\mathcal{F}_{i}^{d'}$}}
        \put(51.2,5){\footnotesize {$\mathcal{M}_i$}}
        \put(75.3,6.9){\footnotesize {$G_{i}^{m}$}}
        \put(74.2,3){\footnotesize {$G_{i}^{gt}$}}
        \put(69,3.2){\footnotesize {$\hat{G}_{i}$}}
        \put(63,2.5){\footnotesize {$\mathcal{F}_{i}^{G}$}}
        \put(70.8,15.6){\footnotesize {$\mathcal{M}_i$}}
        \put(55,15.6){\footnotesize {$\mathcal{A}_{i}^{G}$}}
        \put(48.2,17.7){\footnotesize {$\mathcal{F}_{i-1}^{d}$}}
        \put(6.5,0.7){\footnotesize \textbf{Inward Reference}}
        \put(30.5,0.7){\footnotesize \textbf{Reconstruction Block}}
        \put(62,0.7){\footnotesize \textbf{Outward Reference}}
    \end{overpic}
    \caption{
        \textbf{Pipeline of the proposed bilateral reference blocks.} 
        The source images at the original scale are combined with decoder features as the inward reference and fed into the reconstruction block, where deformable convolutions with hierarchical receptive fields are employed. The aggregated features are then used to predict the gradient maps in the outward reference. Gradient-aware features are then turned into the attention map to act on the original features.
    }
    \label{fig:BiRef}
\end{figure*}

\subsection{Localization Module}
\label{sec:localization_module}
For a batch of HR images $\mathcal{I}\in{}\mathbb{R}^{N \times 3 \times H \times W}$ as input, the transformer encoder~\cite{swin_v1} extracts features at different stages,~\ie{}, $\mathcal{F}_1^e,\mathcal{F}_2^e,\mathcal{F}_3^e,\mathcal{F}^e$ with resolutions as $\{[\frac{H}{k},\frac{W}{k}],k=4,8,16,32\}$. The features of the first four stages $\{\mathcal{F}_i^e\}_{i=1}^3$ are transferred to the corresponding decoder stages with lateral connections (1$\times$1 convolution layers). Meanwhile, they are stacked and concatenated in the last encoder block to generate $\mathcal{F}^e$.

The encoder output feature $\mathcal{F}^e$ is then fed into a classification module, where $\mathcal{F}^e$ is led into a global average pooling layer and a fully connected layer for classification with the category $C$ to obtain a better semantic representation for localization. HR features are squeezed in the bottleneck. To enlarge the receptive fields to cover features of large objects and focus on local features for high precision simultaneously~\cite{denseASPP}, which is important for HR tasks involved, we employ ASPP modules~\cite{deeplab_v3} here for multi-context fusion. $\mathcal{F}^e$ is squeezed to $\mathcal{F}^d$ for transfer to the reconstruction module.

\subsection{Reconstruction Module}
\label{sec:reconstruction_module}
The setting of the receptive field (RF) has been a challenge of HR segmentation. Small RFs lead to inadequate context information to locate the right target on a large background, whereas large RFs often result in insufficient feature extraction in detailed areas. To achieve balance, we propose the reconstruction block (RB) in each BiRef block as a replacement for the vanilla residual blocks. In RB, we employ deformable convolutions~\cite{deformable_conv} with hierarchical receptive fields (\ie{}, 1$\times$1, 3$\times$3, 7$\times$7) and an adaptive average pooling layer to extract features with RFs of various scales. These features extracted by different RFs are then concatenated as $\mathcal{F}_{i}^{\theta}$, followed by a 1$\times$1 convolution layer and a batch normalization layer to generate the output feature of RM $\mathcal{F}_{i}^{d'}$.
In the reconstruction module, the squeezed feature $\mathcal{F}^d$ is fed into the BiRef block for the feature $\mathcal{F}_3^d$.  With $\mathcal{F}_3^l$, the first BiRef block predicts coarse maps, which are then reconstructed into higher-resolution versions through the following BiRef blocks. Following~\cite{FPN}, the output feature of each BiRef block $\mathcal{F}_i^d$ is added with its lateral feature $\mathcal{F}_i^l$ of the LM at each stage,~\ie{}, $\{\mathcal{F}_i^{d+}=\textit{Upsample}\uparrow{}(\mathcal{F}_i^{d} + \mathcal{F}_i^l), i=,3,2,1\}$. Meanwhile, all BiRef blocks generate intermediate predictions $\{\mathcal{M}_i\}_{i=3}^1$ by multi-stage supervision, with resolutions in ascending order. Finally, the last decoding feature $\mathcal{F}_1^{d+}$ is passed through an 1$\times$1 convolution layer to obtain the final predicted maps $\mathcal{M}\in{}\mathbb{R}^{N\times{}1\times{}H\times{}W}$.

\subsection{Bilateral Reference}
\label{sec:bilateral_reference}
In DIS, HR training images are very important for deep models to learn details and perform highly accurate segmentation. However, most segmentation models follow previous works~\cite{UNet,FPN} to design the network architecture in an encoder-decoder structure with down-sampling and up-sampling, respectively. Besides, due to the large size of the input, concentrating on the target objects becomes more challenging. To deal with these two main problems, we propose \emph{bilateral reference}, consisting of an inward reference (InRef) and an outward reference (OutRef), which is illustrated in~\cref{fig:BiRef}. Inward reference and outward reference play the roles of supplementing HR information and drawing attention to areas with dense details, respectively.

In InRef, images $\mathcal{I}$ with original high resolution are cropped to patches $\{\mathcal{P}_{k=1}^N\}$ of consistent size with the output features of the corresponding decoder stage. These patches are stacked with the original feature $\mathcal{F}_i^{d+}$ to be fed into the RM. Existing methods with similar techniques either add $\mathcal{I}$ only at the last decoding stage~\cite{UDUN} or resize $\mathcal{I}$ to make it applicable with original features in low resolution. Our inward reference avoids these two problems through adaptive cropping and supplies the necessary HR information at every stage.

In OutRef, we use gradient labels to draw more attention to areas of richer gradient information, which is essential for the segmentation of fine structures. First, we extract the gradient maps of the input images as $G_{i}^{gt}$. Meanwhile, $\mathcal{F}_{i}^{\theta}$ is used to generate the feature $\mathcal{F}_{i}^{G}$ to produce the predicted gradient maps $\hat{G}_{i}$. With this gradient supervision, $\mathcal{F}_{i}^{G}$ is sensitive to the gradient. It passes through a conv and a sigmoid layer and is used to generate the gradient referring attention $\mathcal{A}_{i}^{G}$, which is then multiplied by $\mathcal{F}_{i}^{d'}$ to generate output of the BiRef block as $\mathcal{F}_{i-1}^{d}$. 

Considering that the background may have non-target noise with a lot of gradient information, we apply a masking strategy to alleviate the influence of non-target areas. We perform morphological operations on intermediate predictions $\mathcal{M}_i$ and use dilated $\mathcal{M}_i$ as a mask. The mask is used to multiply the gradient map $G_{i}^{gt}$ to generate $G_{i}^{m}$, where the gradients outside the mask area are removed.

\subsection{Objective Function}
In HR segmentation tasks, using only pixel-level supervision (BCE loss) usually results in the deterioration of detailed structural information in HR data. Inspired by the great results in~\cite{BASNet} which used a hybrid loss, we use BCE, IoU, SSIM, and CE losses together to collaborate for the supervision on the levels of pixel, region, boundary, and semantic, respectively.
The final objective function is a weighted combination of the above losses and can be formulated as:
\begin{equation}
\label{eqn:loss_final}
    \begin{split}
      {L} &= {L}_\text{pixel} + {L}_\text{region} + {L}_\text{boundary} + {L}_\text{semantic} \\
          &= \lambda_1 {L}_\text{BCE} + \lambda_2 {L}_\text{IoU} + \lambda_3 {L}_\text{SSIM} + \lambda_4 {L}_\text{CE}\;,
    \end{split}
\end{equation}
where $\lambda_1,~\lambda_2,~\lambda_3,$ and $\lambda_4$ are respectively set to 30, 0.5, 10, and 5 to keep all the losses on the same quantitative level at the beginning of the training. The final objective function consists of binary cross-entropy (BCE) loss, intersection over union (IoU) loss, structural similarity index measure (SSIM) loss, and cross-entropy (CE) loss. The complete definition of losses can be found below.

\begin{itemize}

\item{}\textbf{BCE loss:} pixel-aware supervision, which is used for pixel-level supervision for the generation of binary maps:
\begin{equation}
    \centering
    {
    \scriptstyle
    L_{BCE} = -\sum\limits_{(i,j)}[G(i,j)\log(M(i,j))+(1-G(i,j))\log(1-M(i,j))]
    },
    \label{equ:loss_BCE}
\end{equation}
where $G(i,j)$ and $\mathcal{M}(i,j)$ denote the value of the GT and binarized predicted maps, respectively, at pixel $(i,j)$.

\item{}\textbf{IoU loss:} region-aware supervision for the enhancement of binary map predictions:
\begin{equation}
    \centering
    {
    L_{IoU} = 1 - \tfrac{\sum\limits_{r=1}\limits^H\sum\limits_{c=1}\limits^WM(i,j)G(i,j)}{\sum\limits_{r=1}\limits^H\sum\limits_{c=1}\limits^W[M(i,j)+G(i,j)-M(i,j)G(i,j)]}
    }.
    \label{equ:loss_IoU}
\end{equation}

\item{}\textbf{SSIM loss:} boundary-aware supervision to improve the accuracy in boundary parts.
Given GT maps $G$ and predicted maps $\mathcal{M}$, $\mathbf{y}=\{y_j:j=1,...,N^2\}$ and $\mathbf{x}=\{x_j:j=1,...,N^2\}$ represent the pixel values of two corresponding $N\times N$ patches derived from $G$ and $\mathcal{M}$, respectively. $SSIM(\mathbf{x}, \mathbf{y})$ is defined as:
\begin{equation}
    \centering
    L_{SSIM}=1 -  \frac{(2\mu_x\mu_y+C_1)(2\sigma_{xy}+C_2)}{(\mu_x^2+\mu_y^2+C_1)(\sigma_x^2+\sigma_y^2+C_2)},
    \label{equ:loss_ssim}
\end{equation}
where $\mu_x$, $\mu_y$ and $\sigma_x$, $\sigma_y$ are the means and standard deviations of $\mathbf{x}$ and $\mathbf{y}$, respectively, $\sigma_{xy}$ is their covariance. $C$ is used to avoid division by zero. 

\item{}\textbf{CE loss:} semantic-aware supervision, which is used to learn better semantic representation:

\begin{equation}
    \centering
    {
    L_{CE} = -\sum_{c=1}^{N} y_{o,c}\log(p_{o,c})
    },
    \label{equ:loss_CE}
\end{equation}
where $N$ is the number of classes, $y_{o,c}$ states whether class label $c$ is the correction classification for observation $o$, and $p_{o,c}$ denotes the predicted probability that $o$ is of class $c$. 

\end{itemize}

\subsection{Training Strategies Tailored for DIS}
\label{sec:train_strategies}

Due to the high cost of training models on HR data, we have explored training tricks for HR segmentation tasks to improve performance and reduce training costs.

First, we found that our model converges relatively quickly in the localization of targets and the segmentation of rough structures (measured by F-measure~\cite{Fmeasure}, S-measure~\cite{Smeasure}) on DIS5K (\eg{}, 200 epochs). However, the performance in segmenting fine parts is still increasing after very long training (\eg{}, 400 epochs), which is reflected in metrics such as $F_{\beta}^{\omega}$ and $HCE_{\gamma}$.
Second, though long training can easily achieve great results in terms of both structure and edges, it consumes too much computation; we found that multi-stage supervision can dramatically accelerate the learning on segmenting fine details and make the model achieve similar performance as before but with only 30\% training epochs. 
Third, we also found that fine-tuning with only region-level losses can easily improve the binarization of predicted results and those metric scores (\eg{}, $F_{\beta}^{\omega}$, $E_{\phi}^{m}$, $HCE$) that are closer to practical use. 
Finally, we used context feature fusion and image pyramid inputs on the backbone, which are commonly used tricks to process HR images with deep models. In experiments, these two modifications to the backbone achieved a general improvement in DIS and similar HR segmentation tasks. 

As shown in~\cref{tab:supp_epochs_mss}, we show the effectiveness of training epochs and the multi-stage supervision. As the results show, our~\ourmodel{} can achieve relatively good results after 200 epochs of training. Continuous training in 400 epochs can increase a small portion of metrics measuring structural information (\eg{}, $F_{\beta}^{x}$, $S_m$), while bringing a larger improvement in metrics measuring fine details (\eg{}, $HCE_{\gamma}$).

Although simple long training can achieve better results, the improvement is relatively small, concerning its high computational cost on HR data. We investigated the multi-stage supervision (MSS), which is a widely used training strategy used in binary segmentation works~\cite{GICD,DIS5K}. Different from MSS in these works for higher precision, it plays a role in accelerating the training convergence. As the results in~\cref{tab:supp_epochs_mss} show, our~\ourmodel{} trained for 200 epochs with MSS can achieve similar performance with it trained for 400 epochs. MSS successfully cut training time in half and can be used for further HR segmentation tasks for more efficient training.

\begin{table}
    \begin{center}
    \renewcommand{\arraystretch}{1.0}
    \setlength\tabcolsep{1.55mm}
    \footnotesize{}
    \caption{\textbf{Quantitative ablation studies of the proposed multi-stage supervision for acceleration and training epochs.}}
    \label{tab:supp_epochs_mss}
    \begin{tabular}{cc|cccccc}
        \hline
        \multicolumn{2}{c|}{Settings}  & \multicolumn{6}{c}{DIS-VD} \\
        MSS & Epoch \metricsDIS{}
        \\
        \hline
         & 200 &            .875 & .848 & .041 & .886 & .914 & 1207 \\
         & 400 &            .897 & .863 & .036 & .905 & .937 & 1039 \\
        \hline
        \rowcolor{myGray}
        \checkmark & 200 &  .892 & .858 & .037 & .901 & .932 & 1043 \\
        \hline
    \end{tabular}
    \end{center}
\end{table}


\section{Experiments}
\label{sec:experiments}
\subsection{Datasets}
\label{sec:datasets}
\textbf{Training Sets.} 
For DIS, we follow~\cite{DIS5K,FP-DIS,UDUN} to use DIS5K-TR as our training set in experiments. For HRSOD, we follow~\cite{PGNet} to set different combinations of HRSOD, UHRSD, and DUTS as the training set. For COD, we follow~\cite{SINet_v2,FSPNet} to use the concealed samples in CAMO-TR and COD10K-TR as the training set.

\textbf{Test Sets.} 
To obtain a complete evaluation of our~\ourmodel{}, we tested it on all test sets in DIS5K (DIS-TE1, DIS-TE2, DIS-TE3, and DIS-TE4). We also conducted an evaluation of \ourmodel{} on the HRSOD test sets (DAVIS-S~\cite{HRSOD}, HRSOD-TE~\cite{HRSOD}, and UHRSD-TE~\cite{PGNet}) and the COD test sets (CAMO-TE~\cite{CAMO}, COD10K-TE~\cite{SINet_v1}, and NC4K~\cite{SLSR}). Low-resolution SOD test sets (DUTS-TE~\cite{DUTS} and DUT-OMRON~\cite{DUT-OMRON}) are additionally used for supplementary experiments.

\subsection{Evaluation Protocol}
\label{sec:evaluation}
For a comprehensive evaluation, we employ the widely used metrics,~\ie{}, S-measure~\cite{Smeasure} ($S_m$), max/mean/weighted F-measure~\cite{Fmeasure} ($F_{\beta}^{x}/F_{\beta}^{m}/F_{\beta}^{\omega}$), max/mean E-measure~\cite{Emeasure} ($E_{\xi}^{x}/E_{\xi}^{m}$), mean absolute error (MAE), and relax HCE~\cite{DIS5K} ($HCE_{\gamma}$) to evaluate performance. Detailed descriptions of these metrics can be found as follows.

\begin{itemize}

\item{}\textbf{S-measure}~\cite{Smeasure} (structure measure, $S_{\alpha}$) is a structural similarity measurement between a saliency map and its corresponding GT map. Evaluation with $S_{\alpha}$ can be obtained at high speed without binarization. The $S_\alpha$-measure is computed as:
\begin{equation}
  S_{\alpha} = \alpha \cdot S_o + (1 - \alpha) \cdot S_{r},
  \label{eqn:Sm}
\end{equation}
where $S_o$ and $S_r$ denote object-aware and region-aware structural similarity, and $\alpha$ is set to 0.5 by default, as suggested by Fan~\etal{}in~\cite{Smeasure}.

\item{}\textbf{F-measure}~\cite{Fmeasure} ($F_{\beta}$) is designed to evaluate the weighted harmonic mean value of precision and recall. The output of the saliency map is binarized with different thresholds to obtain a set of binary saliency predictions. The predicted saliency maps and GT maps are compared to obtain precision and recall values. F-measure can be computed as:
\begin{equation}
  F_{\beta} = \frac{(1+\beta^2) \cdot Precision \cdot Recall}{\beta^2 \cdot Precision + Recall},
  \label{eqn:Fm}
\end{equation}
where $\beta^2$ is set to 0.3 to emphasize precision over recall, following~\cite{SOD_review1}.  The maximum F-measure score obtained with the best threshold for the entire dataset is used and denoted as $F_{\beta}^{x}$.

\item{}\textbf{E-measure}~\cite{Emeasure} (enhanced-alignment measure, $E_{\xi}$) is designed as a perceptual metric to evaluate the similarity between the predicted maps and the GT maps both locally and globally. 
E-measure is defined as:
\begin{equation}
  E_{\xi} = \frac{1}{W  H}\sum_{x=1}^{W}\sum_{y=1}^{H}\phi_{\xi}(x, y),
  \label{eqn:Em}
\end{equation}
where $\phi_{\xi}$ indicates the enhanced alignment matrix. Similarly to the F-measure, we also adopt the maximum E-measure ($E_{\xi}^{x}$) and also the mean E-measure ($E_{\xi}^{m}$) as our evaluation metrics.

\item{}\textbf{MAE} (mean absolute error, $\epsilon$) is a simple pixel-level evaluation metric that measures the absolute difference between non-binarized predicted results $\mathcal{M}$ and GT maps $G$.
It is defined as:
\begin{equation}
  \epsilon = \frac{1}{W  H}\sum_{x=1}^{W}\sum_{y=1}^{H}|\hat{Y}(x,y) - \text{G}(x, y)|.
  \label{eqn:MAE}
\end{equation}

\item{}\textbf{$\mathbf{HCE_\gamma}$}~\cite{DIS5K} (human correction efforts, $HCE$) is a newly proposed metric aimed at evaluating the human efforts required to correct faulty predictions to satisfy specific accuracy requirements in real-world applications.
Specifically, $HCE$ is quantified by the approximate number of mouse clicks. In practical applications, minor prediction errors can be tolerated. Therefore, relax $HCE$ ($HCE_{\gamma}$) is introduced, where $\gamma$ denotes tolerance. In experiments, we use $HCE_{5}$ (relax $HCE$ with $\gamma = 5$) to stay consistent with that of the original paper~\cite{DIS5K}, where detailed descriptions of $HCE_\gamma$ are provided.

\end{itemize}

\begin{table}
\begin{center}
\renewcommand{\arraystretch}{1.0}
\setlength\tabcolsep{1.0mm}
\footnotesize{}
\caption{\textbf{Quantitative ablation studies of the proposed components in the proposed~\ourmodel{}.} The ablation studies are conducted on the effectiveness of the proposed components, including reconstruction module (RM), inward reference (InRef), outward reference (OutRef), and their combinations.}
\label{tab:ablation_modules}
\begin{tabular}{ccc|cccccc}
\hline
\multicolumn{3}{c|}{Modules}  & \multicolumn{6}{c}{DIS-VD} \\
RM & InRef & OutRef \metricsDIS{}
\\
\hline
  &  &  &                        .837 & .785 & .056 & .845 & .887 & 1204  \\
 \checkmark &  &  &              .855 & .831 & .048 & .865 & .895 & 1167 \\
  & \checkmark &  &              .848 & .825 & .050 & .857 & .903 & 1152 \\
 \checkmark & \checkmark &  &    .869 & .834 & .041 & .886 & .912 & 1093 \\
 \checkmark &  & \checkmark &    .863 & .831 & .042 & .891 & .918 & 1106 \\
  & \checkmark & \checkmark &    .861 & .839 & .044 & .881 & .911 & 1114 \\
\hline
\rowcolor{myGray}
 \checkmark & \checkmark & \checkmark &
                                 .889 & .851 & .038 & .900 & .924 & 1065 \\
\hline
\end{tabular}
\end{center}
\end{table}

\begin{table}
\begin{center}
\renewcommand{\arraystretch}{1.0}
\setlength\tabcolsep{1.2mm}
\footnotesize{}
\caption{\textbf{Effectiveness of practical strategies for training high-resolution segmentation.} The experimental comparison of the proposed several tricks for HR segmentation tasks is provided here, including context feature fusion (CFF), image pyramids input (IPT), regional loss fine-tuning (RLFT), and their combinations. The results are obtained by our final model.}
\label{tab:ablation_tricks}
\begin{tabular}{ccc|cccccc}
\hline
\multicolumn{3}{c|}{Modules}  & \multicolumn{6}{c}{DIS-VD} \\
CFF & IPT & RLFT \metricsDIS{}
\\
\hline
  &  &  &                        .889 & .851 & .038 & .900 & .924 & 1065 \\
 \checkmark &  &  &              .893 & .856 & .038 & .904 & .928 & 1054 \\
  & \checkmark &  &              .895 & .857 & .037 & .904 & .927 & 1051 \\
 & & \checkmark &    .890 & .861 & .036 & .899 & .932 & 1043 \\
\hline
\rowcolor{myGray}
 \checkmark & \checkmark & \checkmark &
                                 .897 &    .863   & .036 &   .905   &  .937  & 1039 \\
\hline
\end{tabular}
\end{center}
\end{table}

\subsection{Implementation Details}
\label{sec:implementation_details}
All images are resized to 1024$\times$1024 for training and testing. The generated segmentation maps are resized (\ie{}, bilinear interpolation) to the original size of the corresponding GT maps for evaluation. Horizontal flip is the only data augmentation used in the training process. The number of categories $C$ is set to 219, as given in DIS-TR. The proposed~\ourmodel{} is trained with Adam optimizer~\cite{Adam} for DIS/HRSOD/COD tasks for 600/150/150 epochs, respectively. The model is fine-tuned with the IoU loss for the last 20 epochs. The initial learning rate is set to $10^{-4}$ and $10^{-5}$ for DIS and others, respectively. Models are trained with PyTorch~\cite{PyTorch} on eight NVIDIA A100 GPUs. The batch size is set to $N$=4 for each GPU during training.

\subsection{Ablation Study}
\label{sec:Ablation}

We study the effectiveness of each component (\ie{}, RM and BiRef) and practical strategies (\ie{}, CFF, IPT, and RLFT) introduced for our~\ourmodel{} and conduct an investigation about their contributions to improved DIS results. Quantitative results regarding each module and strategy are shown in~\cref{tab:ablation_modules} and~\ref{tab:ablation_tricks}, respectively.

\begin{figure}[!t]
    \centering
    \footnotesize{}
    \begin{overpic}[width=.49\textwidth]{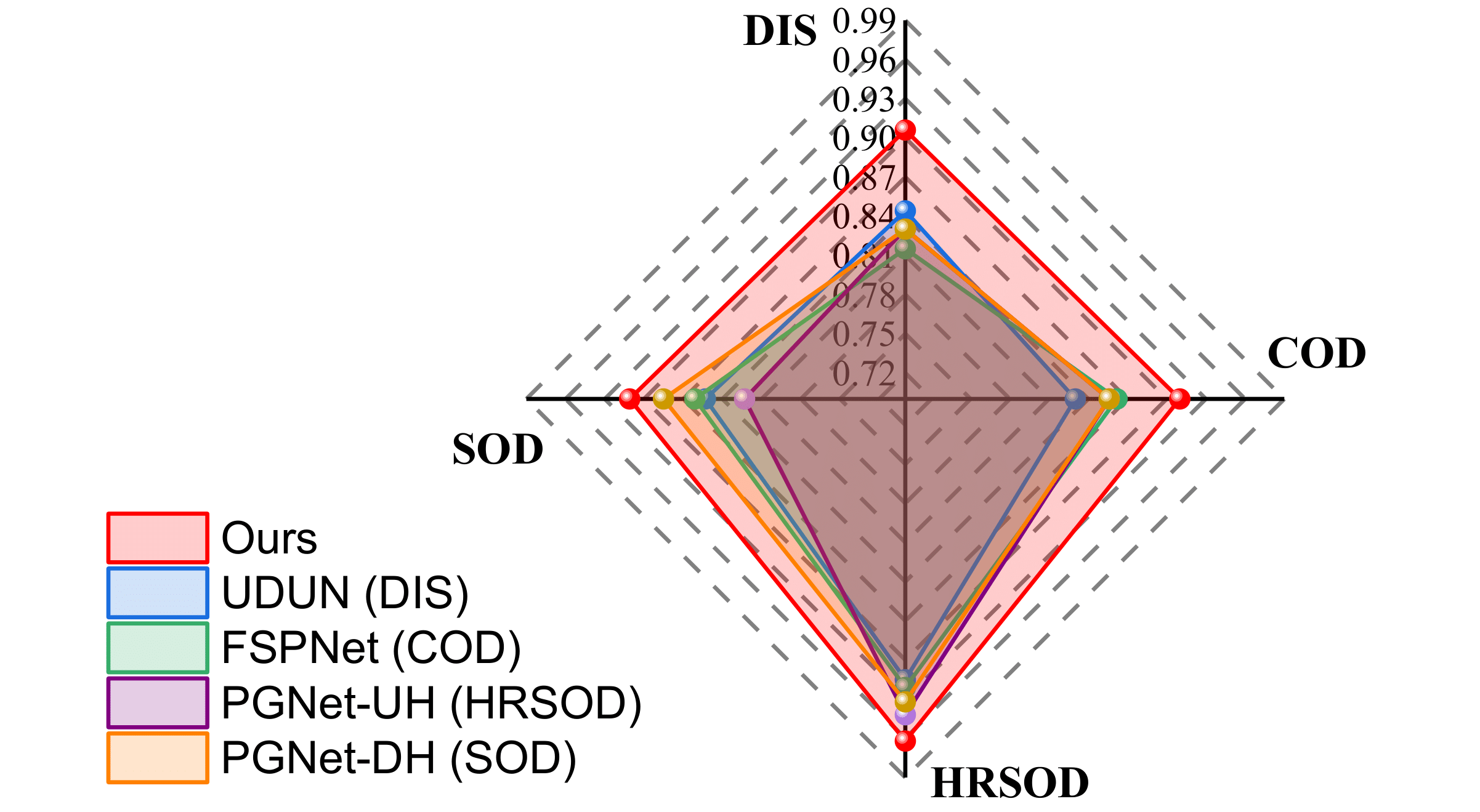}
    \end{overpic}
    \caption{\textbf{Quantitative comparisons of the proposed~\ourmodel{}~and the best task-specific models.} S-measure~\cite{Smeasure} is used for the comparison here. UDUN~\cite{UDUN}, FSPNet~\cite{FSPNet}, PGNet-UH~\cite{PGNet}, and PGNet-DH~\cite{PGNet} are currently the best models for the DIS, COD, HRSOD, and SOD tasks, respectively.}
    \label{fig:sota_radar}
\end{figure}

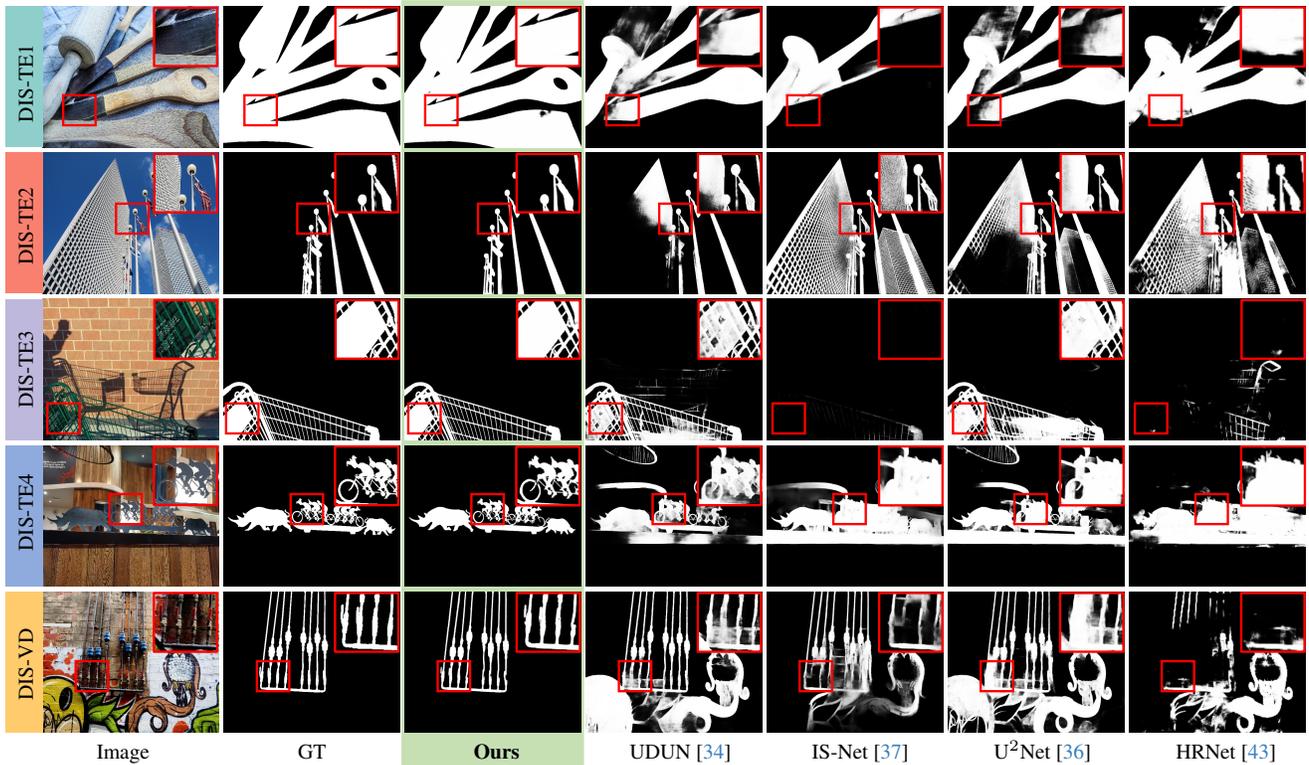
\begin{figure*}[!t]
    \centering
    \footnotesize{}
    \begin{overpic}[width=.99\textwidth]{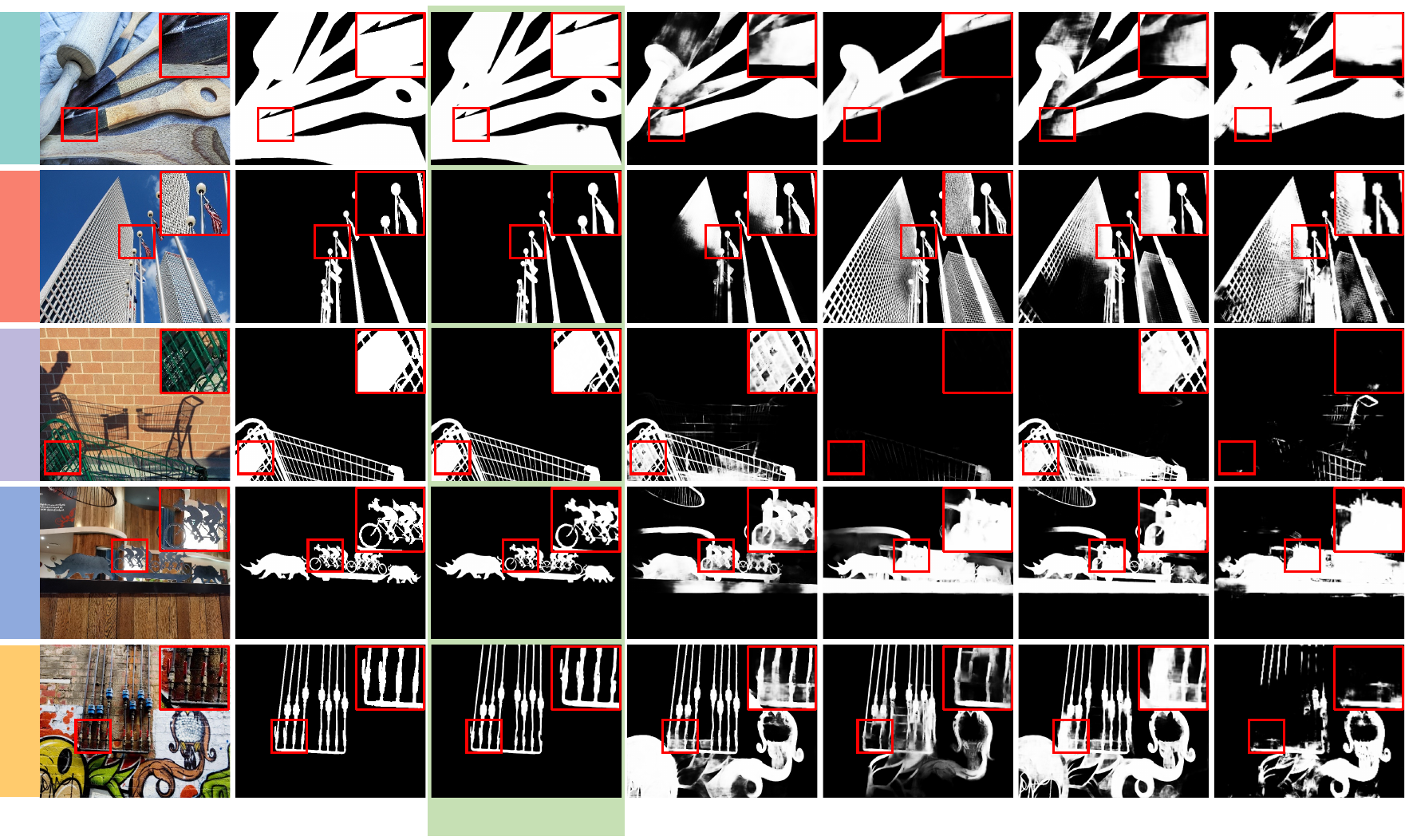}
        \put(7,1){Image}
        \put(22.5,1){GT}
        \put(36,1){\textbf{Ours}}
        \put(48,1){UDUN~\cite{UDUN}}
        \put(62,1){IS-Net~\cite{DIS5K}}
        \put(76,1){U$^2$Net~\cite{U-2-Net}}
        \put(90,1){HRNet~\cite{HRNet}}
        \put(1.0,50){\rotatebox{90}{DIS-TE1}}
        \put(1.0,39){\rotatebox{90}{DIS-TE2}}
        \put(1.0,28){\rotatebox{90}{DIS-TE3}}
        \put(1.0,17){\rotatebox{90}{DIS-TE4}}
        \put(1.0,5.5){\rotatebox{90}{DIS-VD}}
    \end{overpic}
    \caption{\textbf{Qualitative comparisons of the proposed~\ourmodel{}~and previous methods on the DIS5K benchmark.} 
    The results of the previous methods are from~\cite{UDUN}, where all models are trained with images in 1024$\times$1024. Zoom in for a better view.}
    \label{fig:qual_dis}
\end{figure*}

\begin{table*}[t!]
    \setlength{\belowcaptionskip}{0cm}   
    \renewcommand{\arraystretch}{1.0}
    \renewcommand{\tabcolsep}{1.36pt}
    \footnotesize
    \centering
    \caption{\textbf{Quantitative comparisons between our \ourmodel{}~and the state-of-the-art methods on DIS5K.} ``$\uparrow$'' (``$\downarrow$'') means that the higher (lower) is better. We use the results from~\cite{UDUN}, where all methods take 1024$\times$1024 input.}
	\label{tab:sota_dis}
	\begin{tabular}{lcccccc|cccccc|cccccc}
		\toprule
		\multirow{2}{*}{\textbf{Methods}} & \multicolumn{6}{c}{\textbf{DIS-TE1 (500)}}  & \multicolumn{6}{c}{\textbf{DIS-TE2 (500)}}& \multicolumn{6}{c}{\textbf{DIS-TE3 (500)}}\\
		\cmidrule[0.05em](lr){2-7} \cmidrule[0.05em](lr){8-14} \cmidrule[0.05em](lr){14-19} 
		\metricsDIS{}\metricsDIS{}\metricsDIS{} \\
		\midrule
        BASNet$_{19}$~\cite{BASNet} & .663 & .577 & .105 & .741 & .756 & 155 & .738 & .653 & .096 & .781 & .808 & 341 & .790 & .714 & .080 & .816 & .848 & 681 \\
        U$^2$Net$_{20}$~\cite{U-2-Net} & .701 & .601 & .085 & .762 & .783 & 165 & .768 & .676 & .083 & .798 & .825 & 367 & .813 & .721 & .073 & .823 & .856 & 738 \\
        HRNet$_{20}$~\cite{HRNet} & .668 & .579 & .088 & .742 & .797 & 262 & .747 & .664 & .087 & .784 & .840 & 555 & .784 & .700 & .080 & .805 & .869 & 1049 \\
        PGNet$_{22}$~\cite{PGNet} & .754 & .680 & .067 & .800 & .848 & 162 & .807 & .743 & .065 & .833 & .880 & 375 & .843 & .785 & .056 & .844 & .911 & 797 \\
        IS-Net$_{22}$~\cite{DIS5K} & .740 & .662 & .074 & .787 & .820 & 149 & .799 & .728 & .070 & .823 & .858 & 340 & .830 & .758 & .064 & .836 & .883 & 687 \\
        FP-DIS$_{23}$~\cite{FP-DIS} & .784 & .713 & .060 & .821 & .860 & 160 & .827 & .767 & .059 & .845 & .893 & 373 & .868 & .811 & .049 & .871 & .922 & 780 \\
        UDUN$_{23}$~\cite{UDUN} & .784 & .720 & .059 & .817 & .864 & 140 & .829 & .768 & .058 & .843 & .886 & 325 & .865 & .809 & .050 & .865 & .917 & 658 \\
        \rowcolor{myGray}
        \textbf{\ourmodel{}} &  .860 &    .819   & .037 &   .885   &  .911  & 106  & .894 &    .857   & .036 &   .900   &  .930  & 266  & .925 &    .893   & .028 &   .919   &  .955  & 569  \\
        \rowcolor{myGray}
        \textbf{\ourmodel{}$_{SwinB}$} &  .857 &    .819   & .038 &   .884   &  .912  & 110 &  .890 &    .854   & .037 &   .898   &  .930  & 275 &  .919 &    .886   & .030 &   .915   &  .953  & 597  \\
        \rowcolor{myGray}
        \textbf{\ourmodel{}$_{SwinT}$} &  .823 &    .774   & .048 &   .855   &  .887  & 117 &  .862 &    .821   & .046 &   .877   &  .912  & 290 &  .899 &    .860   & .036 &   .897   &  .942  & 627  \\
        \rowcolor{myGray}
        \textbf{\ourmodel{}$_{PVTv2b2}$} &  .839 &    .796   & .042 &   .870   &  .903  & 111 &  .881 &    .842   & .040 &   .888   &  .925  & 280 &  .903 &    .866   & .036 &   .901   &  .941  & 614  \\
		\midrule
		\multirow{2}{*}{\textbf{Methods}} & \multicolumn{6}{c}{\textbf{DIS-TE4 (500)}}  & \multicolumn{6}{c}{\textbf{DIS-TE (1-4) (2,000)}}& \multicolumn{6}{c}{\textbf{DIS-VD (470)}}\\
		\cmidrule[0.05em](lr){2-7} \cmidrule[0.05em](lr){8-14} \cmidrule[0.05em](lr){14-19} 
		\metricsDIS{}\metricsDIS{}\metricsDIS{} \\
        \midrule
        BASNet$_{19}$~\cite{BASNet} & .785 & .713 & .087 & .806 & .844 & 2852 & .744 & .664 & .092 & .786 & .814 & 1007 & .737 & .656 & .094 & .781 & .809 & 1132 \\
        U$^2$Net$_{20}$~\cite{U-2-Net} & .800 & .707 & .085 & .814 & .837 & 2898 & .771 & .676 & .082 & .799 & .825 & 1042 & .753 & .656 & .089 & .785 & .809 & 1139 \\
        HRNet$_{20}$~\cite{HRNet} & .772 & .687 & .092 & .792 & .854 & 3864 & .743 & .658 & .087 & .781 & .840 & 1432 & .726 & .641 & .095 & .767 & .824 & 1560 \\
        PGNet$_{22}$~\cite{PGNet} & .831 & .774 & .065 & .841 & .899 & 3361 & .809 & .746 & .063 & .830 & .885 & 1173 & .798 & .733 & .067 & .824 & .879 & 1326 \\
        IS-Net$_{22}$~\cite{DIS5K} & .827 & .753 & .072 & .830 & .870 & 2888 & .799 & .726 & .070 & .819 & .858 & 1016 & .791 & .717 & .074 & .813 & .856 & 1116 \\
        FP-DIS$_{23}$~\cite{FP-DIS} & .846 & .788 & .061 & .852 & .906 & 3347 & .831 & .770 & .047 & .847 & .895 & 1165 & .823 & .763 & .062 & .843 & .891 & 1309 \\
        UDUN$_{23}$~\cite{UDUN} & .846 & .792 & .059 & .849 & .901 & 2785 & .831 & .772 & .057 & .844 & .892 & 977 & .823 & .763 & .059 & .838 & .892 & 1097 \\
        \rowcolor{myGray}
        \textbf{\ourmodel{}} &  .904 &    .864   & .039 &   .900   &  .939  & 2723 & .896 &    .858   & .035 &   .901   &  .934  & 916  & .891 &    .854   & .038 &   .898   &  .931  & 989  \\
        \rowcolor{myGray}
        \textbf{\ourmodel{}$_{SwinB}$} &  .899 &    .860   & .040 &   .895   &  .938  & 2836 &  .891 &    .855   & .036 &   .898   &  .933  & 954 &  .881 &    .844   & .039 &   .890   &  .925  & 1029  \\
        \rowcolor{myGray}
        \textbf{\ourmodel{}$_{SwinT}$} &  .880 &    .834   & .049 &   .878   &  .925  & 2888 &  .866 &    .822   & .045 &   .877   &  .916  & 980 &  .862 &    .819   & .045 &   .874   &  .917  & 1070  \\
        \rowcolor{myGray}
        \textbf{\ourmodel{}$_{PVTv2b2}$} &  .890 &    .846   & .045 &   .886   &  .929  & 2871 &  .878 &    .838   & .041 &   .886   &  .925  & 969 &  .868 &    .827   & .044 &   .880   &  .919  & 1073  \\
		\bottomrule
	\end{tabular}
\end{table*}

\textbf{Baseline}. We provide a simple but strong encoder-decoder network as the baseline for the DIS task. To capture better hierarchical features on various scales, we chose the Swin transformer large~\cite{swin_v1} as our default backbone network. Then, to obtain a better semantic representation in the DIS task, we divided the images in DIS-TR into 219 classes according to their label names and added an auxiliary classification head at the end of the encoder. In the decoder of the baseline network, each decoder block is made up of two residual blocks~\cite{resnet}. All stages of the encoder and decoder are connected with an 1$\times{}$1 convolution, except the deepest stage, where an ASPP~\cite{deeplab_v3} block is used for connectivity. With this setup, our baseline network has outperformed existing DIS models in most metrics, as shown in~\cref{tab:ablation_modules} and~\cref{tab:sota_dis}.

\textbf{Reconstruction Module.} As shown in~\cref{tab:ablation_modules}, our model gains an overall improvement with the proposed RM. The RM provides multi-scale receptive fields on the HR features for local details and overall semantics. It brings $\sim$2.2\% $F_{\beta}^x$ relative improvement with little extra computational cost.

\textbf{Bilateral Reference.} We separately investigate the effectiveness of the inward reference (InRef, with source images) and the outward reference (OutRef, with gradient labels) in BiRef. InRef supplemented lossless HR information globally, while OutRef drew more attention to the fine-detail parts to achieve higher precision in those areas. As shown in~\cref{tab:ablation_modules}, they work jointly to bring 2.9\% $F_{\beta}^x$ relative improvement to \ourmodel{}. RM and BiRef are combined to achieve 6.2\% $F_{\beta}^x$ relative improvement.

\textbf{Training Strategies.} As shown in~\cref{tab:ablation_tricks}, the proposed strategies improve performance from different perspectives. CCF and IPT improve overall performance, while RLFT specifically improves precision in edge details, which is reflected in metrics such as $F_{\beta}^{\omega}$ and $HCE_{\gamma}$.

\subsection{State-of-the-Art Comparison}
\label{sec:competing_methods}
To validate the general applicability of our method, we conduct extensive experiments on four tasks,~\ie{}, high-resolution dichotomous image segmentation (DIS), high-resolution salient object detection (HRSOD), concealed object detection (COD), and salient object detection (SOD). We compare our proposed~\ourmodel{} with all the latest task-specific models on existing benchmarks~\cite{DIS5K,HRSOD,PGNet,DUTS,DUT-OMRON,CAMO,SINet_v2,SLSR}.

\begin{table*}[t!]
    \setlength{\belowcaptionskip}{0cm}   
    \renewcommand{\arraystretch}{1.0}
    \renewcommand{\tabcolsep}{1.1pt}
    \footnotesize
    \centering
    \caption{\textbf{Quantitative comparisons between our \ourmodel{}~and the state-of-the-art methods in high-resolution and low-resolution SOD datasets.} TR denotes the training set. To provide a fair comparison, we train our~\ourmodel{} with different combinations of training sets, where 1, 2, and 3 represent DUTS~\cite{DUTS}, HRSOD~\cite{HRSOD}, and UHRSD~\cite{PGNet}, respectively.}
    \label{tab:sota_sod}
    \vspace{-7pt}
    \begin{tabular}{lccccc|cccc|cccc|cccc|cccc}
	\toprule
	\multicolumn{2}{c}{\multirow{3}{*}{\textbf{Test Sets}}} & \multicolumn{12}{c|}{\textbf{High-Resolution Benchmarks}} & \multicolumn{8}{c}{\textbf{Low-Resolution Benchmarks}}\\
	\cmidrule[0.05em](lr){3-22}  
	\multirow{3}{*}{\textbf{Methods}} & \multirow{3}{*}{\textbf{TR}} & \multicolumn{4}{c}{\textbf{DAVIS-S (92)}}  & \multicolumn{4}{c}{\textbf{HRSOD-TE (400)}}& \multicolumn{4}{c}{\textbf{UHRSD-TE (988)}} & \multicolumn{4}{c}{\textbf{DUTS-TE (5,019)}} & \multicolumn{4}{c}{\textbf{DUT-OMRON(5,168)}}\\
	\cmidrule[0.05em](lr){3-6} \cmidrule[0.05em](lr){7-10} \cmidrule[0.05em](lr){11-14} \cmidrule[0.05em](lr){15-18} \cmidrule[0.05em](lr){19-22}  
	&\metricsSODHR{}\metricsSODHR{}\metricsSODHR{}\metricsSODLR{}\metricsSODLR{} \\
	\midrule
        LDF$_{20}$~\cite{LDF} & 1           & .922 & .911 & .947 & .019 & .904 & .904 & .919 & .032 & .888 & .913 & .891 & .047 & .892 & .898 & .910 & .034 & .838 & .820 & .873 & .051 \\
        HRSOD$_{19}$~\cite{HRSOD} & 1,2     & .876 & .899 & .955 & .026 & .896 & .905 & .934 & .030 & - & - & - & - & .824 & .835 & .885 & .050 & .762 & .743 & .831 & .065 \\
        DHQ$_{21}$~\cite{DHQSOD} & 1,2      & .920 & .938 & .947 & .012 & .920 & .922 & .947 & .022 & .900 & .911 & .905 & .039 & .894 & .900 & .919 & .031 & .836 & .820 & .873 & .045 \\
        PGNet$_{22}$~\cite{PGNet} & 1       & .935 & .936 & .947 & .015 & .930 & .931 & .944 & .021 & .912 & .931 & .904 & .037 & .911 & .917 & .922 & .027 & .855 & .835 & .887 & .045 \\
        PGNet$_{22}$~\cite{PGNet} & 1,2     & .948 & .950 & .975 & .012 & .935 & .937 & .946 & .020 & .912 & .935 & .905 & .036 & .912 & .919 & .925 & .028 & .858 & .835 & .887 & .046 \\
        PGNet$_{22}$~\cite{PGNet} & 2,3     & .954 & .957 & .979 & .010 & .938 & .945 & .946 & .020 & .935 & .949 & .916 & .026 & .859 & .871 & .897 & .038 & .786 & .772 & .884 & .058 \\
        \rowcolor{myGray}
        \textbf{\ourmodel{}} & 1          & .967   &  .966 &  .984  & .008 & .957   &  .958 &  .972  & .014 & .931   &  .933 &  .943  & .030 & .939   &  .937 &  .958  & .019 & .868   &  .813 &  .878  & .040  \\
        \rowcolor{myGray}
        \textbf{\ourmodel{}} & 1,2        & .973   &  .976 &  .990  & .006 & .962   &  .963 &  .976  & .011 & .937   &  .942 &  .951  & .024 & .938   &  .935 &  .960  & .018 & .868   &  .818 &  .882  & .040  \\
        \rowcolor{myGray}
        \textbf{\ourmodel{}} & 1,3        & .975   &  .977 &  .989  & .006 & .959   &  .958 &  .972  & .014 & .952   &  .960 &  .965  & .019 & .942   &  .942 &  .961  & .018 & .881   &  .837 &  .896  & .036  \\
        \rowcolor{myGray}
        \textbf{\ourmodel{}} & 2,3        & .976   &  .980 &  .990  & .006 & .956   &  .953 &  .967  & .016 & .952   &  .958 &  .964  & .019 & .933   &  .928 &  .954  & .020 & .864   &  .810 &  .879  & .040  \\
        \rowcolor{myGray}
        \textbf{\ourmodel{}} & 1,2,3      & .975   &  .979 &  .989  & .006 & .962   &  .961 &  .973  & .013 & .957   &  .963 &  .969  & .016 & .944   &  .943 &  .962  & .018 & .882   &  .839 &  .896  & .038  \\
	\bottomrule
	\end{tabular}
\end{table*}

\begin{table*}[t!]
    \setlength{\belowcaptionskip}{0cm}   
    \renewcommand{\arraystretch}{1.0}
    \renewcommand{\tabcolsep}{2.4pt}
    \footnotesize
    \centering
    \vspace{-5pt}
    \caption{\textbf{Comparison of \ourmodel{} with recent methods.} As seen, \ourmodel{} performs much better than previous methods.}\label{tab:sota_cod}
    \vspace{-7pt}
    \begin{tabular}{lcccccc|cccccc|cccccc}
        \toprule
        \multirow{2}{*}{\textbf{Methods}} & \multicolumn{6}{c}{\textbf{CAMO (250)}}  & \multicolumn{6}{c}{\textbf{COD10K (2,026)}}& \multicolumn{6}{c}{\textbf{NC4K (4,121)}}\\
        \cmidrule[0.05em](lr){2-7} \cmidrule[0.05em](lr){8-13} \cmidrule[0.05em](lr){14-19} 
        \metricsCOD{}
        \metricsCOD{}
        \metricsCOD{} \\
        \midrule
		SINet$_{20}$~\cite{SINet_v1}   & .751 & .606 & .675 & .771 & .831 & .100 & .771 & .551 & .634 & .806 & .868 & .051 & .808 & .723 & .769 & .871 & .883 & .058 \\
        BGNet$_{22}$~\cite{BGNet}   & .812 & .749 & .789 & .870 & .882 & .073 & .831 & .722 & .753 & .901 & .911 & .033 & .851 & .788 & .820 & .907 & .916 & .044 \\
        SegMaR$_{22}$~\cite{SegMaR}  & .815 & .753 & .795 & .874 & .884 & .071 & .833 & .724 & .757 & .899 & .906 & .034 & .841     & .781     & .820     & .896     & .907   & .046     \\
        ZoomNet$_{22}$~\cite{ZoomNet} & .820 & .752 & .794 & .878 & .892 & .066 & .838 & .729 & .766 & .888 & .911 & .029 & .853 & .784 & .818 & .896 & .912 & .043 \\
        SINetv2$_{22}$~\cite{SINet_v2} & .820 & .743 & .782 & .882 & .895 & .070 & .815 & .680 & .718 & .887 & .906 & .037 & .847 & .770 & .805 & .903 & .914 & .048 \\
        FEDER$_{23}$~\cite{FEDER} & .802   &    .738   &  .781  &  .867  &  .873 & .071 &   .822   &    .716   &  .751  &  .900  &  .905 & .032 &   .847   &    .789   &  .824  &  .907  &  .915 & .044 \\
        HitNet$_{23}$~\cite{HitNet}& .849 & .809 & .831 & .906 & .910 & .055 & .871 & .806 & .823 & .935 & .938 & .023 & .875 & .834 & .853 & .926 & .929 & .037 \\
        FSPNet$_{23}$~\cite{FSPNet}& .856 & .799 & .830 & .899 & .928 & .050 & .851 & .735 & .769 & .895 & .930 & .026 & .879 & .816 & .843 & .915 & .937 & .035 \\
        \rowcolor{myGray}
        \textbf{\ourmodel{}}    & .904   &    .890   &  .904 &  .954  &  .959 & .030 & .913   &    .874   &  .888  &  .960  &  .967 & .014 & .914   &    .894   &  .909  &  .953  &  .960 & .023 \\
        \bottomrule
    \end{tabular}
    \vspace{-5pt}
\end{table*}

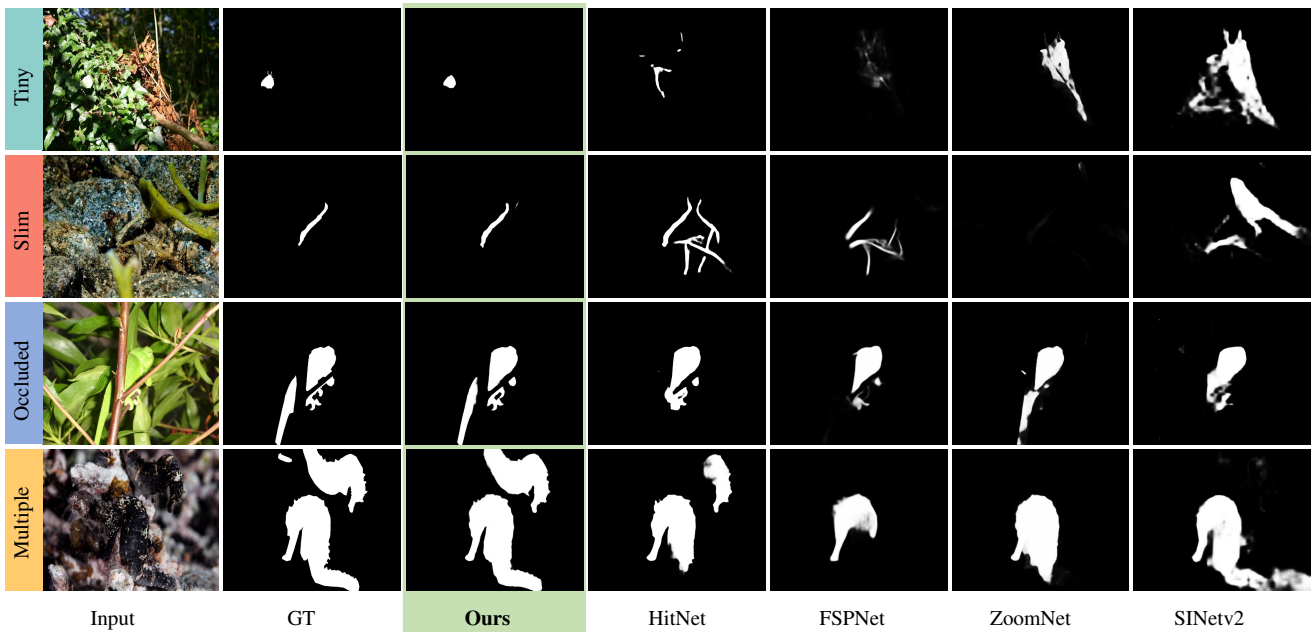
\begin{figure*}[t!]
    \centering
    \footnotesize
    \begin{overpic}[width=\textwidth]{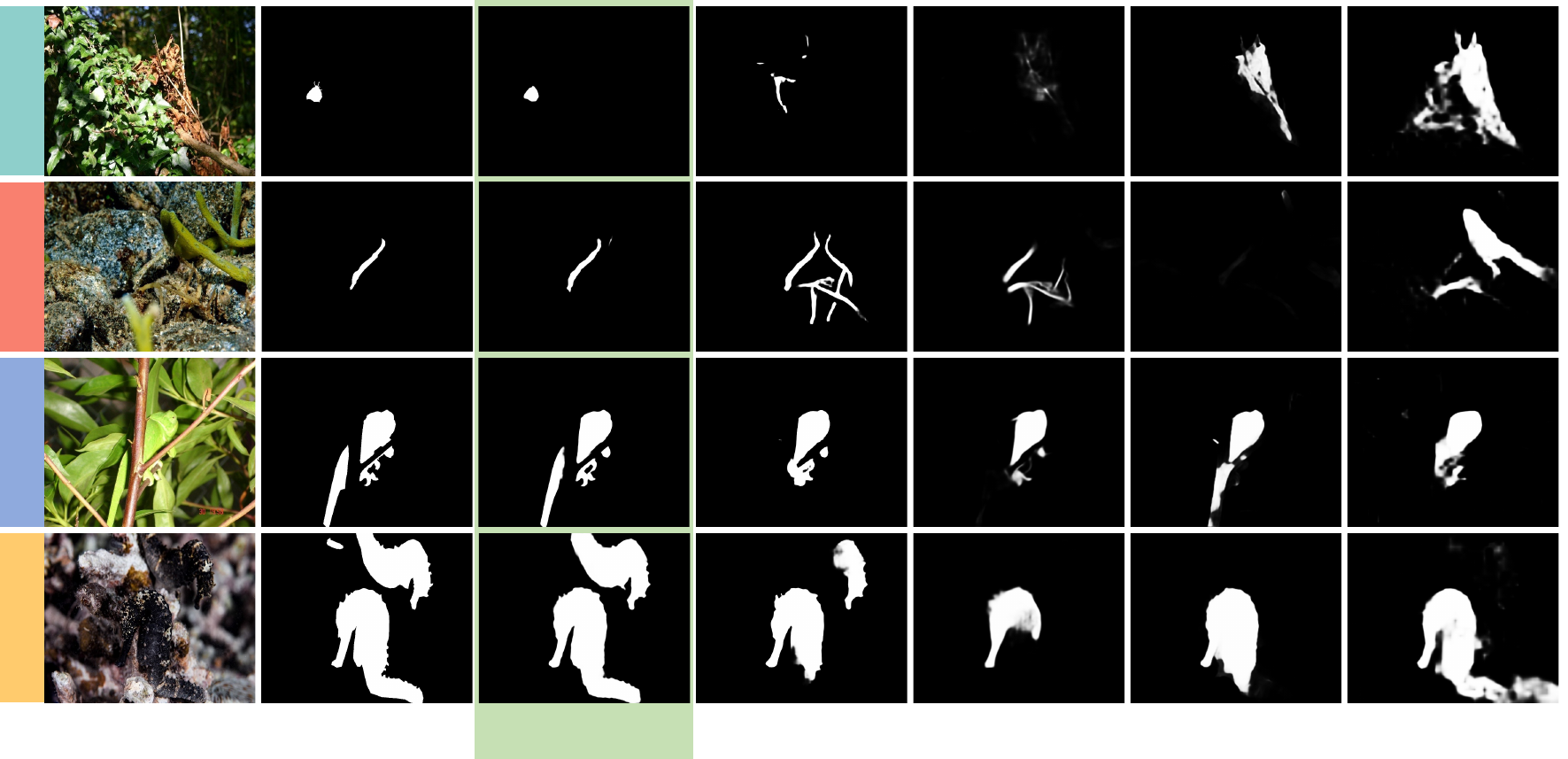}
        \put(0.7,40.5){\rotatebox{90}{Tiny}}
        \put(0.7,29.5){\rotatebox{90}{Slim}}
        \put(0.7,16.5){\rotatebox{90}{Occluded}}
        \put(0.7,6){\rotatebox{90}{Multiple}}
        \put(6.5,1){Input}
        \put(21.5,1){GT}
        \put(35,1){\textbf{Ours}}
        \put(49,1){HitNet}
        \put(62,1){FSPNet}
        \put(75,1){ZoomNet}
        \put(89,1){SINetv2}
    \end{overpic}
    \vspace{-15pt}
    \caption{\textbf{Visual comparisons of the proposed~\ourmodel{}~and other competitors on COD10K benchmark.} Samples with different challenges are provided here to show the superiority of~\ourmodel{} from different perspectives.}
    \vspace{-10pt}
    \label{fig:qual_cod}
\end{figure*}

\begin{table*}[t!]
    \footnotesize
    \centering
    \setlength{\belowcaptionskip}{0cm}   
    \renewcommand{\arraystretch}{1.0}
    \renewcommand{\tabcolsep}{22.3pt}
    \caption{Comparison of different DIS methods on the performance, efficiency, and model complexity. Full details can be referred to~\url{https://drive.google.com/drive/u/0/folders/1s2Xe0cjq-2ctnJBR24563yMSCOu4CcxM}.}
    \label{tab:acc_eff}
    \vspace{-5pt}
    \begin{tabular}{l|c|c|c|c}
        \toprule
        \textbf{Model}  & \textbf{Runtime (ms)} & \textbf{\#Params (MB)} & \textbf{MACs (G)} & \textbf{DIS-TEs} ($HCE, F_{\beta}^{\omega}$) \\
    \midrule
        \ourmodel{}$_{SwinL}$       & 83.3 & 215    & 1143  & 916, .858 \\
    \midrule
        \ourmodel{}$_{SwinL\_cp}$   & 78.3 & 215    & 1143  & 916, .858 \\
    \midrule
        \ourmodel{}$_{SwinB}$       & 61.4 & 101    & 561   & 954, .855 \\
    \midrule
        \ourmodel{}$_{SwinT}$       & 40.9 & 39     & 231   & 980, .822 \\
    \midrule
        \ourmodel{}$_{PVTv2b2}$     & 47.8 & 35     & 195   & 969, .838 \\
    \midrule
        \ourmodel{}$_{PVTv2b1}$     & 36.6 & 23     & 147   & 978, .817 \\
    \midrule
        \ourmodel{}$_{PVTv2b0}$     & 32.9 & 11     & 89    & 1013, .806 \\
    \midrule
        IS-Net                      & 16.0 & 44     & 160   & 1016, .726 \\
    \midrule
        UDUN$_{Res50}$              & 33.5 & 25     & 142   & 977, .772 \\
    \bottomrule
    \end{tabular}
\end{table*}

\begin{figure}[t!]
    \footnotesize
    \centering
    \begin{overpic}[width=\linewidth]{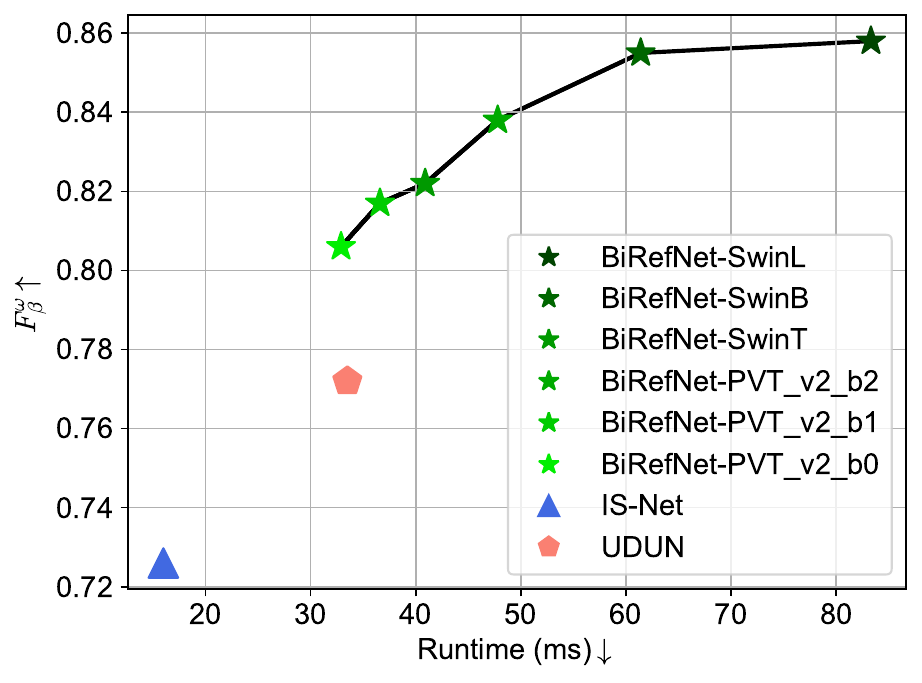}
    \end{overpic}
    \vspace{-10pt}
    \caption{Comparison of the efficiency, size, complexity, and performance of~\ourmodel{} and existing DIS methods.}
    \label{fig:acc_eff}
    \vspace{-10pt}
\end{figure}

\textbf{Quantitative Results.}~\cref{tab:sota_dis} shows a quantitative comparison between the proposed~\ourmodel{} and previous state-of-the-art methods. Our~\ourmodel{} outperforms all previous methods in widely used metrics. The complexities of DIS-TE1$\sim$DIS-TE4 are in ascending order. The metrics for structure similarity (\eg{}, $S_{\alpha}$, $E_{\phi}^{x}$) focus more on global information. Pixel-level metrics, such as MAE ($M$), emphasize the precision of details. Metrics based on mean values (\eg{}, $E_{\phi}^{m}$, $F_{\phi}^{m}$) better match the requirements of practical applications where maps are thresholded. As seen in~\cref{tab:sota_dis}, our~\ourmodel{} outperforms previous methods not only on the accuracy of the global shape but also in the details of the pixels. It is noteworthy that the results are better, especially in metrics that cater more to practical applications.

Additionally, our~\ourmodel{} outperforms existing task-specific models on the HRSOD and COD tasks. As shown in~\cref{tab:sota_sod}, \ourmodel{} achieved much higher accuracy on both high-resolution and low-resolution SOD benchmarks. Compared with the previous SOTA method~\cite{PGNet}, our~\ourmodel{} achieved an average improvement of 2.0\% $S_m$. Furthermore, as shown in~\cref{tab:sota_cod}, in the COD task, \ourmodel{} also shows a much better performance compared to the previous SOTA models, with an average improvement of 5.6\% $S_m$ on the three widely used COD benchmarks. These results show the remarkable generalization ability of our~\ourmodel{} to similar HR tasks. 

For a clearer illustration of the generalizability and powerful performance of~\ourmodel{}, we provide a radar picture shown in~\cref{fig:sota_radar}, where we run our model and the best task-specific models on DIS/HRSOD/COD/SOD tasks. As the results show, our~\ourmodel{} achieves leading results in all four tasks. The other task-specific models show their weakness in similar HR segmentation tasks. For example, FSPNet~\cite{FSPNet} ranks second in COD benchmarks, while it ranks fourth/third/third in DIS/HRSOD/SOD tasks, respectively.

\textbf{Qualitative Results.}~\cref{fig:qual_dis} shows segmentation maps produced by the most competitive existing DIS models and the proposed~\ourmodel{}. As the results show, we provide samples of all test sets and one validation set. \ourmodel{} outperforms the previous DIS methods from two perspectives,~\ie{}, the location of target objects and the more accurate segmentation of the details of the objects.
For example, in the samples of DIS-TE4 and DIS-TE2, there are neighboring distractors that attract the attention of other models to produce false positives. On the contrary, our~\ourmodel{} eliminates the distractors and accurately segments the target. In samples of DIS-TE3 and DIS-VD, \ourmodel{} shows much greater performance in precisely segmenting areas where fine details are rich. Compared with previous methods, our~\ourmodel{} can clearly segment slim shapes and curved edges.

We also provide a qualitative comparison on the COD task.~\cref{fig:qual_cod} shows hard samples with different challenges. For example, in the row of the occluded frog, the area of the frog is divided by the branch that covers it, while our~\ourmodel{} can accurately segment the scattered fragments almost the same as the GT map. In contrast, in the results of the other methods, fragments are difficult to find all, let alone to provide precise segmentation maps. For tiny and slim objects, \ourmodel{} shows a better ability to find the right target. Our~\ourmodel{} also shows superiority in finding multiple concealed objects.

\begin{figure*}[t!]
    \centering
    \footnotesize
    \begin{overpic}[width=\textwidth]{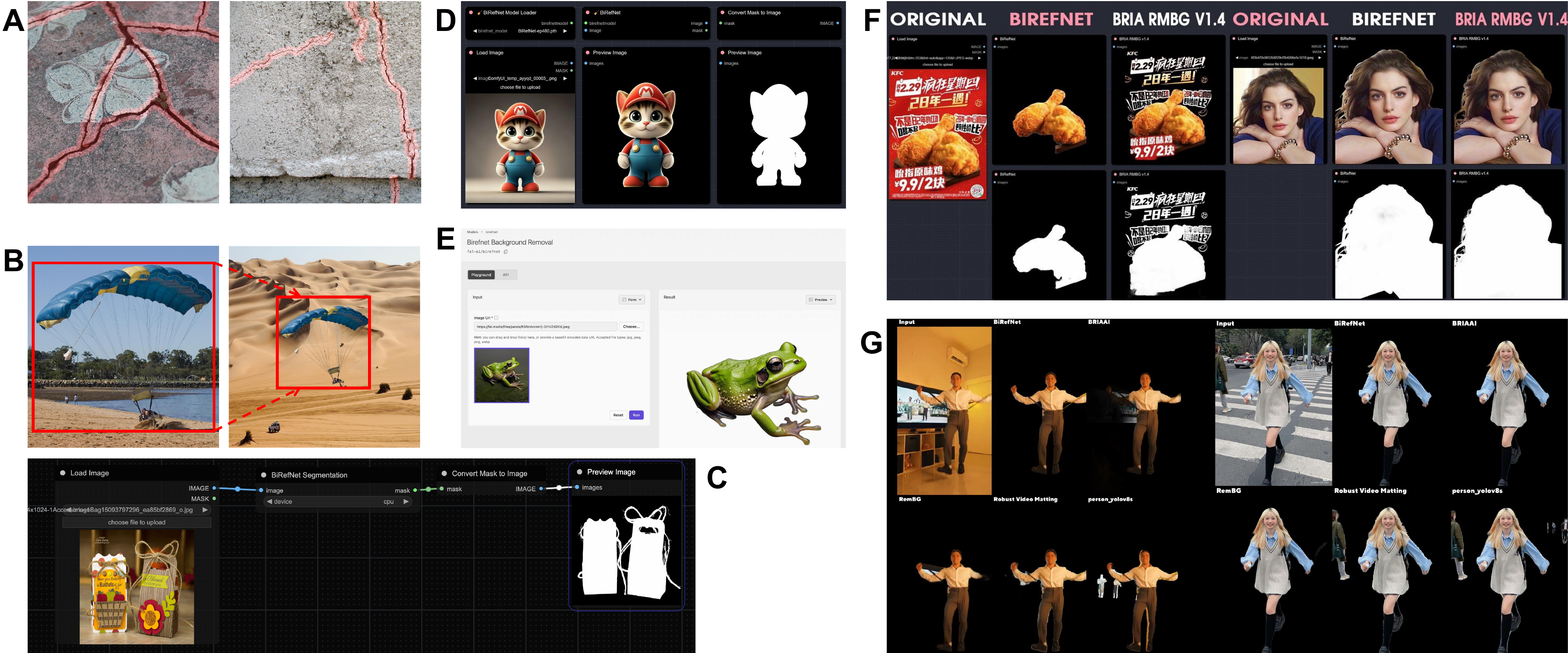}
    \end{overpic}
    \caption{\textbf{Potenial applications and selected existing third-party applications based on~\ourmodel{}, and visual comparisons on social media.} (A) Potential application \#1. Building crack detection for the maintenance of architecture health. (B) Potential application \#2. Highly accurate object extraction in high-resolution natural images. (C) A project by viperyl first packs our~\ourmodel{} as a ComfyUI node and makes this SOTA model easier to use for everyone. (D) ZHO also provides a ComfyUI-based project to further improve the UI for our~\ourmodel{}, especially for video data. (E) Fal.AI encapsulates our~\ourmodel{} online with more useful options in UI and API to call the model. (F) ZHO provides a visual comparison between our~\ourmodel{} and previous SOTA method BRIA RMGB v1.4 which has extra training on their private training dataset. (G) Toyxyz conducted a comparison between our~\ourmodel{} and previous competitive human matting methods (\eg{}, BRIAAI, RemBG, Robust Video Matting, and Person YOLOv8s) with both videos and images.}
    \label{fig:applications}
\end{figure*}

\textbf{Efficiency and Complexity Comparison.} We equip our~\ourmodel{} with different backbones to obtain models in different sizes. Runtime, number of parameters, MACs, and performance of them are further tested to provide a comprehensive comparison between them and other methods. First, we provide a quantitative comparison in~\cref{tab:acc_eff}. The FPS of the largest BiRefNet can be more than 10, which is acceptable in most practical applications. We also used the compiled version (\ourmodel{}$_{SwinL\_cp}$) by PyTorch 2.0~\cite{PyTorch} on~\ourmodel{}$_{SwinL}$ to accelerate its inference by 13\%. In addition, we draw the performance and runtime of each model in~\cref{fig:acc_eff} for a clearer display.~\ourmodel{} with different backbones are evaluated and compared with existing DIS methods on DIS-TEs and DIS-VD. Different methods are drawn in different colors and markers. All tests are conducted on a single NVIDIA A100 GPU and an AMD EPYC 7J13 CPU.

\section{Potential Applications}
\label{sec:potential_applications}

We envisage that generated fine segmentation maps have the potential to be utilized in various practical applications.

\textbf{Potential Application \#1 Crack Detection.}
The quality of the walls is important for the health of the architecture~\cite{app_crack}. However, segmentation models trained on commonly used datasets (\eg{}, COCO~\cite{COCO}) can only segment regular foreground objects. The proposed~\ourmodel{} trained on the DIS5K dataset is more aware of the fine details and can also segment targets with higher shape complexities. As shown in~\cref{fig:applications} (A), our~\ourmodel{} can accurately find cracks in the walls and help maintain when to repair them.

\textbf{Potential Application \#2 Highly Accurate Object Extraction.}
Foreground object extraction and background removal have been popular applications in recent years. However, commonly seen methods fail to generate high-quality results when target objects have too high shape complexities~\cite{U-2-Net,BASNet} or need manual guidance (\eg{}, scribble, point, and coarse mask) for more accurate segmentation~\cite{SOD4ImageMatting1,SOD4ImageMatting2}. The proposed~\ourmodel{} trained on DIS5K can generate results with much higher resolution and segment thin threads at the hair level without a mask, as shown in~\cref{fig:applications} (B). On the basis of such refined results, there may be numerous successful downstream applications in the future.

\section{Third-Party Creations}
\label{sec:third-party_creations}
Since the release of our project on Mar 7, 2024, it has attracted much attention from many researchers and developers in the community to promote it spontaneously. Furthermore, great third-party applications have also been made based on our~\ourmodel{}. Due to the rapid growth of relevant works, we only list some typical ones.

\textbf{\#1 Practical Applications.}
Because of the excellent performance of our~\ourmodel{}, more and more third-party applications have been created by developers in the community\footnote{\url{https://github.com/comfyanonymous/ComfyUI}}\footnote{\url{https://github.com/ZHO-ZHO-ZHO/ComfyUI-BiRefNet-ZHO}}. As shown in~\cref{fig:applications} (C and D), some developers have integrated our~\ourmodel{} into the ComfyUI as a node, which helps a lot in matting foreground segmentation to better processing in the subsequent stable diffusion models. For better online access, Fal.AI has established an online demo of our~\ourmodel{} running on an A6000 GPU\footnote{\url{https://fal.ai/models/birefnet}}\footnote{Thanks to FAL.AI for providing us with additional computation resources for further explorations on more practical applications.}, as shown in~\cref{fig:applications} (E). In addition to the common prediction of results, this online application also provides an API service for easy use with HTTP requests.

\textbf{\#2 Social Media.}
In recent days, our~\ourmodel{} has drawn attention from the community. Many tweets have been posted on the X platform (formerly Twitter)\footnote{\url{https://twitter.com/search?q=birefnet&src=typed_query}}. ZHO provides a visual comparison between our~\ourmodel{} and other methods, as given in~\cref{fig:applications} (F).~\ourmodel{} achieves competitive results with the previous SOTA method BRIA RMGB v1.4 in their tests\footnote{\url{https://twitter.com/ZHOZHO672070/status/1771026516388041038}}. It should be noted that our~\ourmodel{} was trained on the training set of the open-source dataset DIS5K~\cite{DIS5K} under MIT license, while the other one was trained on their carefully selected private data and cannot be used for commercial use. As shown in~\cref{fig:applications} (G), more comparisons on both video and image data have been provided by Toyxyz on X between our~\ourmodel{} and previous great foreground human matting methods\footnote{\url{https://twitter.com/toyxyz3/status/1771413245267746952}}. In addition to these posts, PurzBeats has also made an animation with our~\ourmodel{} and uploaded relevant videos\footnote{\url{https://twitter.com/i/status/1772323682934775896}}. A video tutorial can also be found on YouTube by `AI is in wonderland' in Japanese about how to use our~\ourmodel{} in ComfyUI\footnote{\url{https://www.youtube.com/watch?v=o2_nMDUYk6s}}.

\section{Conclusions}
\label{sec:conclusions}
This work proposes a \ourmodel{} framework equipped with a bilateral reference, which can perform dichotomous image segmentation, high-resolution (HR) salient object detection, and concealed object detection in the same framework. With the comprehensive experiments conducted, we find that unscaled source images and a focus on regions of rich information are vital to generating fine and detailed areas in HR images. To this end, we propose the bilateral reference to fill in the missing information in the fine parts (inward reference) and guide the model to focus more on regions with richer details (outward reference). This significantly improves the model's ability to capture tiny-pixel features. To alleviate the high training cost of HR data training, we also provide various practical tricks to deliver higher-quality prediction and faster convergence. Competitive results on 13 benchmarks demonstrate outstanding performance and strong generalization ability of our~\ourmodel{}. We also show that the techniques of~\ourmodel{} can be transferred and used in many practical applications. We hope that the proposed framework can encourage the development of unified models for various tasks in the academic community and that our model can empower and inspire the developer community to create more great works.

{
    \small
    \bibliographystyle{utils/ieee_fullname}
    \bibliography{main}
}


\end{document}